\documentclass{article} 
\usepackage{gram2026, times}

\usepackage{amsmath,amsfonts,bm}

\def\pz{{\phantom 0}}

\def\eqref#1{equation~\ref{#1}}

\def\1{\bm{1}}

\DeclareMathAlphabet{\mathsfit}{\encodingdefault}{\sfdefault}{m}{sl}
\SetMathAlphabet{\mathsfit}{bold}{\encodingdefault}{\sfdefault}{bx}{n}

\usepackage{tikz}
\usepackage{booktabs}       
\usepackage{amsfonts}       
\usepackage{nicefrac}       
\usepackage{microtype}      
\usepackage{xcolor}         
\usepackage{xspace}
\usepackage{amsmath}
\usepackage{etoc}

\newcommand\sd[1]{{\scriptsize{(#1)}}}

\usepackage{multirow}

\def\mnist{MNIST\xspace}

\def\medmnist{MedMNIST3D\xspace}
\def\smoke{SMOKE\xspace}

\newcommand\MSE{\ensuremath{\mathrm{MSE}}}
\newcommand\REG{\ensuremath{\mathrm{REG}}}
\newcommand\ALG{\ensuremath{\mathrm{ALG}}}

\usepackage{mathtools}

\newcommand\iveq{\ensuremath{\mathrm{(\textit{iv})}}\xspace}
\usepackage{wrapfig}
\usepackage{subcaption}
\newcommand{\ignore}[1]{}

\def\K{\ensuremath{\mathbb{K}\xspace}}

\usepackage{amsthm}
\usepackage{calc}
\usepackage{array}

\theoremstyle{plain}
\newtheorem{theorem}{Theorem}

\newtheorem{lemma}[theorem]{Lemma}
\newtheorem{proposition}[theorem]{Proposition}

\theoremstyle{definition}
\newtheorem{definition}[theorem]{Definition}

\usepackage{makecell} 
\usepackage{enumitem}

\usepackage[most]{tcolorbox}
\tcbuselibrary{listingsutf8}

\tcbset{
  highlightbox/.style={
    colback=gray!10,
    colframe=black!30,
    boxrule=0.5pt,
    arc=2pt,
    left=4pt,
    right=4pt,
    top=4pt,
    bottom=4pt,
    enhanced,
    sharp corners,
  }
}

\newcounter{commcounter}
\usepackage{hyperref}
\usepackage{url}

\title{Algebraic Priors for Approximately \\Equivariant Networks}

\author{Riccardo Ali, Pietro Li\`{o} \& Jamie Vicary\\
University of Cambridge\\
\texttt{\{rma55,pl219,jv258\}@cam.ac.uk} \\
}

\def\newtimes{\!\times\!}

\iclrfinalcopy 
\maintrack 
\begin{document}
\etocdepthtag.toc{main}

\maketitle

\begin{abstract}
Equivariant neural networks incorporate symmetries through group actions, embedding them as an inductive bias to improve performance. Existing methods learn an equivariant action on the latent space, or design architectures that are equivariant by construction. These approaches often deliver strong empirical results but can involve architecture-specific constraints, large parameter counts, and high computational cost. We challenge the paradigm of complex equivariant architectures with a parameter-free approach grounded in group representation theory. We prove that for an equivariant encoder over a finite group, the latent space must almost surely contain one copy of its regular representation for each linearly independent data orbit, which we explore with a number of empirical studies. Leveraging this foundational algebraic insight, we impose the group's regular representation as an inductive bias via an auxiliary loss, adding no learnable parameters. Our extensive evaluation shows that this method matches or outperforms specialized models in several cases, even those for infinite groups. We further validate our choice of the regular representation through an ablation study, showing it consistently outperforms defining and trivial group representation baselines.
\end{abstract}

\section{Introduction}

\begin{wrapfigure}{r}{0.5\textwidth}
\centering
\vspace{-20pt}
\[
\begin{tikzpicture}[yscale=.45, xscale=1]
\path [use as bounding box] (-1,-2.2) rectangle +(5,5.2);
\draw [thick, blue!80!black, -latex] (-1,1.5) -- +(1,0);
\draw [thick, blue!80!black, -latex] (1,1.5) -- +(1,0);
\draw [thick, blue!80!black, -latex] (3,1.5) -- +(1,0);
\draw [thick, blue!80!black, fill=blue!20] (0,0) -- ++(0,3) -- ++(1,-.8) -- ++(0,-1.4) -- cycle;
\draw [thick, blue!80!black, fill=blue!20] (3,0) -- ++(0,3) -- ++(-1,-.8) -- ++(0,-1.4) -- cycle;
\node at (2.5,1.5) {$D$};
\node at (0.5,1.5) {$E$};
\node (I) at (-.5,.5) {$\mathcal X$};
\node (O) at (3.5,.5) {$\mathcal Y$};
\draw [->] (-.65,-.2) to [out=-100, in=-80, looseness=9] +(0.2,0);
\draw [->] (3.35,-.2) to [out=-100, in=-80, looseness=9] +(0.2,0);
\node (I) at (-.55,-1.4) {$\alpha_{\mathcal X}$};
\node (I) at (3.45,-1.4) {$\alpha_{\mathcal Y}$};
\draw [->, ] (1.37,-.2) to [out=-100, in=-80, looseness=9] +(0.2,0);
\node [] (I) at (1.47,-1.4) {$\rho_{Z}$};
\node [] (L) at (1.5,.5) {$Z$};
\end{tikzpicture}
\]

\caption{Generic architecture with input set $\mathcal X$, latent space $Z$ and output set $\mathcal Y$, with group actions $\alpha_{\mathcal X}$, $\alpha_{\mathcal Y}$ on the input and output sets, and potentially a representation $\rho_{Z}$ on the latent space.\label{fig:arch}}\vspace{-10pt}

\end{wrapfigure}
\color{black}
Equivariance is a powerful inductive bias used to align model representations with the underlying group structure of the data~\citep{bronstein2021geometricdeeplearninggrids}, as shown in Figure~\ref{fig:arch}. Strict equivariance assumes a level of symmetry that real-world data rarely possesses~\citep{holmes2012turbulence, holl2020phiflow, Yang_2023}, thus over-constraining architectures can degrade their performance~\citep{petrache2025approximationgeneralizationtradeoffsapproximategroup}. \textit{Approximate equivariance} offers a mechanism to better balance geometric structure with representational flexibility~\citep{wang2021equivariant,NEURIPS2021_fc394e99}.

Augmented architectures achieve approximate equivariance by modifying the model structure at the cost of increased computational complexity and often tied to specific architectures \citep{Veefkind2024}. 
Avoiding such design constraints, loss-based approaches enforce equivariance via soft constraints as an additional penalty term~\citep{dupont2020equivariantneuralrendering,koyamaNeuralFourierTransform2024a, jin2024learning}. While flexible, these loss-based methods often rely on arbitrary priors or heuristics, leaving the factors determining the learned representations' geometry  underexplored or confined to specific empirical settings \citep{Bokman_2024}.

In this work, we provide a theoretical and empirical analysis of soft-constrained equivariance for linear representations of finite groups under data augmentations. We analytically prove that, in this setup, the latent space \textit{must contain the regular representation almost surely}, and we give a lower bound of its multiplicity. Our empirical studies, validating the theory, prove the regular representation a principled inductive bias.
We contrast this choice against both learnable and other fixed representations, showing competitive or improved performance. 
This yields a simple, scalable, and model-agnostic framework achieving strong results on invariant and approximately-equivariant benchmarks. We summarize our contributions as follows:

\begin{itemize}[leftmargin=*, nosep]
\item We formally characterize latent space representations, proving that for an approximately equivariant encoder, the regular representation must emerge almost surely. We derive a lower bound on its multiplicity and empirically validate that neural networks tend to learn a linear representation aligned with this structure.
\item We demonstrate that enforcing the regular representation yields a lightweight framework that matches or exceeds state-of-the-art performance. Crucially, our method requires no additional learnable parameters and only a single hyperparameter, whereas competing baselines often demand $5\text{--}20\times$ more parameters.
\item We empirically establish the optimality of the regular representation across a diverse suite of invariant and equivariant tasks. We validate this choice by contrasting it against both learnable approaches and alternative fixed representations, such as the trivial and defining representations.

\end{itemize}


\color{black}
\section{Related Work}
\label{sec:relatedwork}

We provide a brief overview of methods most relevant to approximate equivariance here; an extended review of the broader literature is available in Appendix~\ref{Appendix:extended-related-work}.

\textbf{Augmented architectures.} While strict equivariance is well-established through steerable CNNs~\citep{cohen2016steerablecnns, e2cnn}, recent work focuses on relaxing these constraints to improve flexibility. Residual Pathway Priors (\textbf{RPP})~\citep{NEURIPS2021_fc394e99} combine a strictly equivariant layer with an unconstrained one, while Lift Expansion (\textbf{LIFT})~\citep{wang2021equivariant} factorizes inputs into equivariant and non-equivariant subspaces. More recently, Probabilistic Steerable CNNs (\textbf{PSCNN})~\citep{Veefkind2024} propose learning the optimal equivariance strength per layer. However, these architectural relaxations often incur significant computational costs, requiring elevated parameter counts to match the performance of simpler baselines (see Section~\ref{sec:experiments}).

\textbf{Loss-based approaches.} An alternative paradigm enforces equivariance via the loss function on the latent space. \cite{dupont2020equivariantneuralrendering} propose fixing the latent representation for neural rendering; however, their approach relies on the defining representation of $O(3)$ and assumes the latent geometry matches the input, limiting its generality. Other methods attempt to learn the transformation, such as Neural Fourier Transforms (\textbf{NFT})~\citep{koyamaNeuralFourierTransform2024a} or subspace steerers~\citep{Bokman_2024}. 


We distinguish our work by identifying the \textit{regular representation} of a finite group as the theoretically and empirically justified fixed target. This eliminates the need for additional learnable parameters or complex architectures. Empirically, this algebraic prior proves robust, frequently outperforming unconstrained baselines and even architectures specifically adapted for continuous symmetries.
\section{Background on group representations}
\label{sec:background}

We review essential aspects of group representation theory for our work. We consider a finite group $G$ and work over a base field $\mathbb K$, assumed to be $\mathbb R$ or $\mathbb C$. The results presented are standard, for which we recommend canonical texts such as~\cite{Fulton2004} and~\cite{James2001}. A glossary of notation and further background on group actions are available in Appendix \ref{Appendix:notation} and \ref{Appendix:group-actions}.

\textbf{Regular representation.} For the case of a finite group, the \textit{regular representation} $\rho_\mathrm{reg}$ is defined as the linearisation of the action of $G$\ on itself. Explicitly, we first define $\K[G]$ as having elements given by  linear combinations of group elements $\smash{\sum_i c_i g_i}$ weighted by coefficients $c_i \in \K$. Then $\rho_\mathrm{reg}$ is defined as a representation on $\K[G]$ as follows:
$
\rho_\mathrm{reg}(g)(\textstyle \sum_i c_i g_i) = \sum_i c_i (gg_i).
$
By construction we have $\mathrm{dim}(\rho_\mathrm{reg}) = |G|$, the size of the group.

\textbf{Irreducibility.} Given representations $\rho_1$ on $V_1$ and $\rho_2$ on $V_2$, their \textit{direct sum} $\rho_1 \oplus \rho_2$ acts on $V_1 \oplus V_2$ component-wise: $(\rho_1 \oplus \rho_2)(g)(v_1, v_2) = (\rho_1(g)v_1, \rho_2(g)v_2)$. The \textit{dimension} of a representation is the dimension of its underlying vector space.
A representation is \textit{irreducible} (or an \textit{irrep}) if it cannot be decomposed into such a sum of non-trivial subrepresentations. For finite groups, any representation $\rho$ decomposes into a direct sum of irreps: $\rho \cong \bigoplus_i m_i \rho_i$, where $m_i$ is the multiplicity.
Over $\mathbb{C}$, the regular representation decomposes as $\rho_\mathrm{reg} \cong \bigoplus_i d_i \rho_i$, containing every irrep $\rho_i$ with multiplicity equal to its dimension $d_i$. For example, the group $S_3$ has just the trivial (dim 1), sign (dim 1) and standard (dim 2) irreps (with the same for $D_3$ as they are isomorphic groups); and the cyclic group $C_n$ has $n$ irreducible representations (all dim 1) over $\mathbb{C}$, one for each $n$th root of unity.
While the decomposition over $\mathbb{R}$ may yield different multiplicities, any real representation can be analyzed via its complexification (for instance, through eigendecomposition). 

\textbf{Orthogonality of representations.}
The presence and multiplicity of an irrep $\rho'$ within a  representation $\rho$ is determined by the inner product of their characters $\chi_\rho(g) = \mathrm{Tr}(\rho(g))$, which is defined as $\langle \chi_\rho, \chi_{\rho'} \rangle = \frac{1}{|G|} \sum_{g \in G} \overline{\chi_\rho(g)} \chi_{\rho'}(g)$. This formula and complexification allow us to decompose any representation into its irreducible components. 

\section{Identifying optimal representations}\label{sec:optimal-rep}
\color{black}
We consider the architecture in Figure~\ref{fig:arch} with data $(x, y) \in {\mathcal X} \times {\mathcal Y}$ and a finite group $G$ acting on input/output spaces via $\alpha_{\mathcal X}, \alpha_{\mathcal Y}$. Let $E_\theta:\mathcal X\to Z \subseteq \mathbb R^d$ be an encoder parameterized by $\theta\in\Theta\subseteq\mathbb R^p$. If $G$ acts on $Z$ via a representation $\rho_Z$, we say $E_\theta$ is \textit{$(\rho_Z,\varepsilon)$-equivariant} if $\sup_{x, g}\|E(\alpha_{\mathcal X}(g)(x))-\rho_Z(g)E(x)\|\leq\varepsilon$, encoding the degree of equivariance of $E_\theta$.
We aim to answer the following research questions (RQs):
\begin{enumerate}[label=\textbf{RQ\arabic*}, nosep, leftmargin=*]
    \item Are there algebraic or geometric constraints for what representations can approximately equivariant models instantiate in the latent space?
    \item What representations do models tend to learn in practice, when free to learn one?
    \item Does explicitly enforcing this emerging structure yield practical benefits?
\end{enumerate}
\color{black}
\begin{figure}[t]
    \centering
    \includegraphics[width=0.85\linewidth]{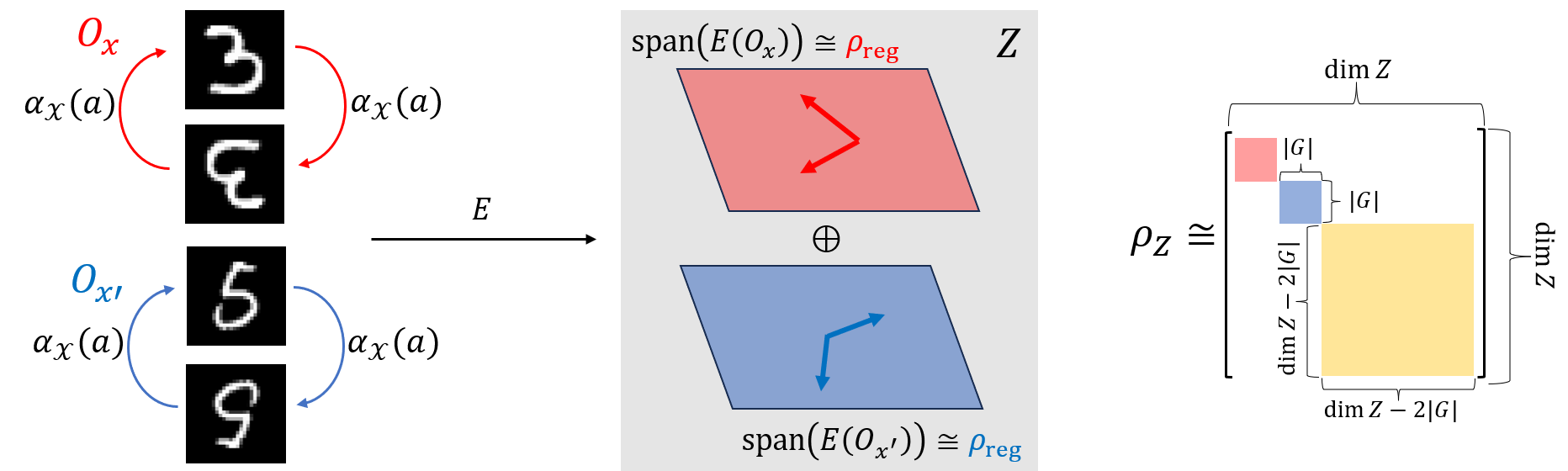}
    \caption{Illustration of our theory for an approximately equivariant encoder $E_\theta$ and $G=C_2=\{1,a\}$, with $\alpha_\mathcal X$ acting by horizontal flips. If $E(\mathcal O_x)$, $E(\mathcal O_{x'})$ are full rank and linearly independent, $Z$ must contain a separate copy of the regular representation $\rho_{\text{reg}}$ for each with probability 1.}
    \label{fig:theory}
    \vspace{-10pt}
\end{figure}

\subsection{Algebraic constraints for approximate equivariance (\textbf{RQ1})}\label{sec:theory}
Let ${\mathcal O}_x:=\{\alpha_{\mathcal X}(g)(x)\,|\,g\in G\}$ denote the \textit{data orbit} of a sample $x$, containing all $G$-augmented versions of $x$. We suppose $x$ is a data sample such that all augmented versions are distinct, i.e. such that $\alpha_{\mathcal X}(g)(x) = \alpha_{\mathcal X}(h)(x)$ implies $g=h$, which is typical for data augmentation.
As a consequence  $|{\mathcal O}_x|=|G|$, and we conclude that $G$ acts freely and transitively on ${\mathcal O}_x$~\citep{nlab}.
We focus on the \textit{embedded orbit} $E_\theta(\mathcal O_x)$. When $\dim(Z) \geq |G|$, we say $E_\theta(\mathcal O_x)$ is \textit{full rank} if its span has dimension $|G|$, corresponding to a strictly positive smallest singular value $\sigma_{\text{min}} > 0$, and \textit{rank deficient} otherwise. We now present our main theoretical results (proofs in Appendix~\ref{Appendix:proofs}).
\color{black}
\begin{theorem}\label{thm:reg-rep}
Let $E_\theta$ be a $(\rho_Z,\varepsilon)$-equivariant encoder and $\sigma_{\text{min}}$ the smallest singular value of $E_\theta(\mathcal O_x)$. 
If $\sigma_{\min} > 0$, there exists a representation $(V,\tilde\rho)$ with $V\subseteq Z$ and $\tilde\rho\cong\rho_{\text{reg}}$ such that for all $g\in G$:
\vspace{-7pt}
\begin{center}
    $||\rho_Z(g)|_V -\tilde\rho(g)||_{\text{op}}\leq  \frac{\varepsilon}{\sigma_{\text{min}}}\sqrt{|G|}$\ \ 
\end{center}
\end{theorem}

\begin{theorem}[informal]
\label{thm:analytic}
If $E_\theta$ is also real analytic in its inputs and parameters, and trained by gradient descent, then for each training sample $x \in \mathcal X$, exactly one of the following holds:
\begin{itemize}
\vspace{-8pt}
\item[(i)] for all possible parameterisations $\theta \in \Theta$, the vectors  $E_\theta(\mathcal O_x)$ are rank deficient, $\sigma_{\text{min}}=0$.
\vspace{-5pt}
\item[(ii)] with probability 1, the vectors $E_\theta(\mathcal O_x)$ are full rank, $\sigma_{\text{min}}>0$. Hence the latent space contains the regular representation, and the bound from Theorem \ref{thm:reg-rep} applies to $E_\theta$.
\end{itemize}
\end{theorem}
\vspace{-3pt}

Analyticity is detailed in Appendix \ref{appendix:analyticity-condition}. 
Case {\em(i)} is restrictive, appearing when $E_\theta$ is structurally constrained (e.g., invariant by design). In contrast, Case {\em(ii)} is generic: for standard architectures, sampling $\theta$ yields a full-rank orbit $E_\theta(\mathcal O_x)$ with probability 1, confirming the presence of the regular representation. Importantly, this property extends across the dataset, where \textit{each linearly independent orbit contributes a separate copy of $\rho_{\text{reg}}$} (see Appendix \ref{Appendix:proofs}).

\textbf{Implications of Theorem \ref{thm:analytic}.} This result highlights a fundamental {tension} in learning approximate symmetries. Consider an encoder trained for invariance via a penalty $\|E(x)-E(\alpha(g)x)\|$. While the optimization objective drives the encoder toward rank deficiency (true invariance), Theorem~\ref{thm:analytic} guarantees that the trajectory remains almost surely in the full-rank regime ($\sigma_{\text{min}} > 0$). Consequently, the network cannot achieve true information erasure; it is forced instead into a state of {signal compression}, where the regular representation is preserved but condensed ($\sigma_{\min} \ll 1$). A {critical question} is whether this theoretical persistence is merely a numerical artifact or representationally significant. To answer this, we employ \textit{linear probes}~\citep{alain2018understandingintermediatelayersusing}, a standard technique in interpretability literature for quantifying feature decodability, in a controlled setup (Rotated MNIST $C4$). Despite $E$ being regularized for invariance, these probes recover the input rotation with 60.6\% accuracy (vs. 25\% chance). See Appendix \ref{Appendix:probes} for more details. This confirms that the attenuated signal is not just numerical noise, but a distinguishing feature that persists despite the training objective. It underscores our core insight: the network does not structurally transition to the trivial representation; rather, it maintains the  algebraic structure of the regular representation, instantiating the target geometry solely through geometric contraction.

\color{black}
\color{black}%
\begin{tcolorbox}[highlightbox]
\textbf{Key theoretical insight:} 
To achieve encoder equivariance in the presence of data augmentation, a sufficiently large latent space must contain a separate copy of the regular representation for each linearly independent full rank embedded orbit. This is summarized in Figure \ref{fig:theory}.
\end{tcolorbox}
\color{black}
The question remains how many copies of the regular representation one obtains in practice, and we
investigate this with the following empirical studies.

\subsection{Empirical exploration of learned representations (\textbf{RQ2})}\label{sec:empirical-exploration}

For our empirical investigation, we conduct experiments with the following loss function:

\begin{align*}
&L_\mathrm{opt} = L_\mathrm{task}\big(D(E(x_i)),\, y_i\big) 
    && \text{{\bf Task loss.} Trains encoder and decoder on the}
    \\[-4pt]
    &&&\text{supervised objective.}
\\[-1pt]
&\quad {}+ \lambda_{t}\, L_\mathrm{task}\!\big(D(\widehat\rho_Z(g)E(x_i)),\, \alpha_{\mathcal Y}(g)(y_i)\big)
&& \text{\textbf{Latent→Output Equivariance.} Encourages}
\\[-4pt]&&&
\text{\(D(\widehat\rho_Z(g)E(x))\) to match \(\alpha_{\mathcal Y}(g)(y)\).}
\\[-1pt]
&\quad {}+ \lambda_{e}\, \mathrm{MSE}\!\big(\widehat\rho_Z(g)E(x_i),\, E(\alpha_{\mathcal X}(g)(x_i))\big)
&& \text{{\bf Input→Latent Equivariance.} Encourages}
\\[-4pt]&&&
\text{\(\widehat\rho_Z(g)E(x)\) to match \(E(\alpha_{\mathcal X}(g)x).\)}
\\[-1pt]
&\quad {}+ \lambda_{a}\big(\mathrm{ALG}_{G,d} + \mathrm{REG}_{G,d}\big)
    && \text{\textbf{Algebra Loss.} Encourages algebraic}
\\[-4pt]&&&
\text{properties for $\widehat \rho_Z$ to be a group representation.}    
\end{align*}
\color{black}

We give additional insight into the algebra loss in Appendix~\ref{Appendix:algebra-loss}. 
We further investigate the latent structure through exploratory studies. Theorems \ref{thm:reg-rep} and \ref{thm:analytic} imply that the learned representation contains copies of the regular representation, with a multiplicity lower-bounded by the number of linearly independent embedded data orbits. We empirically assess this multiplicity in controlled environments, ensuring coverage of invariant versus equivariant objectives, abelian and non-abelian groups ($C_2, C_4, D_3$), and geometric versus non-geometric actions. We observe that analytic encoders (initialized via $\mathcal N(\bm 0,\bm 1)$) consistently converge to a representation composed \textit{entirely} of linearly independent copies of the regular representation (methodology detailed in Appendix \ref{appendix:extract-orbits}). Further investigations into the algebra loss, alternative depths, and initialization schemes are provided in Appendices \ref{Appendix:algebra-loss}, \ref{appendix:exploratory-different-depths}, and \ref{appendix:tmnist-initialisations}.

\begin{table}[h]
\def\m{\text{-}}
\small
\setlength{\tabcolsep}{1mm}
\caption{\textcolor{black}{Left (Right) TMNIST (MNIST) analytic autoencoder task, learned representations of $C_2$ ($D_3$) on the  latent space. $Z$ is taken as the middle layer, which carries the output of the encoder.}\label{tab:tmnist}}
\vspace{-5pt}
    \centering
    \hspace{-1cm}
    \begin{tabular}{cccccc}
    \toprule
    &\multicolumn{2}{c}{Irrep. counts}
    \\
    \cmidrule{2-3}
        Run & \hspace{5pt}$-1$ & $+1$ & Alg.\ loss & Eq. loss & Orbs
        \\ \midrule
        1 & \hspace{5pt}3 & 5 & $9.9 \newtimes 10^{\m 10}$ & $1.6 \newtimes 10^{\m 3}$ & 3 \\
        2 & \hspace{5pt}4 & 4 & $1.1 \newtimes 10^{\m 7\pz}$ & $1.3 \newtimes 10^{\m 3}$ & 4\\
        3 & \hspace{5pt}3 & 5 & $7.4 \newtimes 10^{\m 10}$ & $1.2 \newtimes 10^{\m 4}$ & 3\\
        4 & \hspace{5pt}4 & 4 & $2.3 \newtimes 10^{\m 10}$ & $9.1 \newtimes 10^{\m 5}$ & 4\\
        5 & \hspace{5pt}4 & 4 & $2.9 \newtimes 10^{\m 9\pz}$ & $1.5 \newtimes 10^{\m 3}$ & 4
        \\ \bottomrule
    \end{tabular}
    \hspace{5pt}
    \begin{tabular}{ccccccc}
        
    \toprule
    &\multicolumn{3}{c}{Irrep. counts}
    \\
    \cmidrule{2-4}
        Run & Triv & Sgn & Std & Alg.\ loss & Eq.\ loss & Orbs.
        \\ \midrule
        1 & 2.98 & 3.1 & 5.98 & $1.1\newtimes 10^{\m 4}$ & $1.4\newtimes 10^{\m 2}$ & 3 \\
        2 & 3.03 & 2.98 & 6.01 & $5.8\newtimes 10^{\m 3}$ & $2.1\newtimes 10^{\m 3}$ & 3 \\
        3 & 3.1 & 2.97 & 5.70 & $1.6\newtimes 10^{\m 4}$ & $2.7\newtimes 10^{\m 2}$ & 3 \\
        4 & 2.96 & 3.15 & 5.75 & $2.3\newtimes 10^{\m 3}$ & $3.1\newtimes 10^{\m 3}$ & 3 \\
        5 & 2.91 & 2.99 & 6.02 & $7.5\newtimes 10^{\m 2}$ & $2.2\newtimes 10^{\m 2}$ & 3 \\ \bottomrule
    \end{tabular}
    \hspace{-1cm}
    \vspace{-10pt}
\end{table}

\textbf{Non-analytic encoders.} While Theorem \ref{thm:analytic} formally assumes analyticity, modern architectures can employ piecewise-analytic activations like ReLU. Theoretically, such functions can be arbitrarily well-approximated by analytic maps (Stone-Weierstrass), suggesting our results should transfer (see discussion in Appendix \ref{appendix:analyticity-condition}). We empirically validate this in Appendix~\ref{appendix:piecewise-analytic-exp}, demonstrating that ReLU-based networks exhibit behavior identical to their analytic counterparts, consistently converging to linearly independent copies of the regular representation.

\paragraph{CNN Equivariant task, abelian group, non-geometric action.}
\vspace{10pt}
\newlength{\figwidth}     
\setlength{\figwidth}{0.13\textwidth} 
\newlength{\figgap}       
\setlength{\figgap}{3pt}  
\newlength{\wrapw}
\setlength{\wrapw}{\dimexpr 2\figwidth + \figgap\relax}
\begin{wrapfigure}{r}{\wrapw} 
  \centering
  \begin{minipage}{\wrapw} 
    \centering
    \setlength{\tabcolsep}{0pt}    
    \renewcommand{\arraystretch}{1} 
    \begin{tabular}{@{}c@{\hspace{\figgap}}c@{}}
      \begin{minipage}{\figwidth}\centering
        \includegraphics[width=\figwidth]{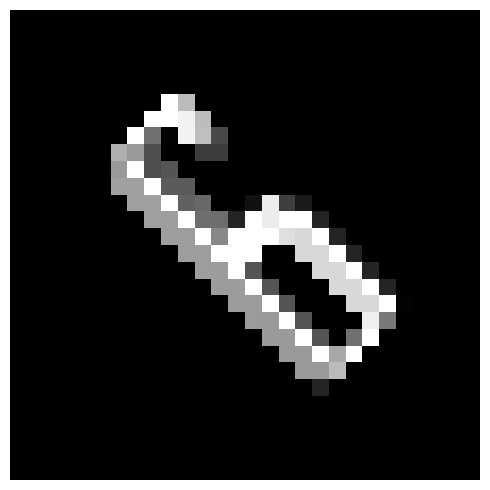}\\[-4pt]
        {\vspace{-2pt}\footnotesize$\scriptstyle x$}
      \end{minipage}
      &
      \begin{minipage}{\figwidth}\centering
        \includegraphics[width=\figwidth]{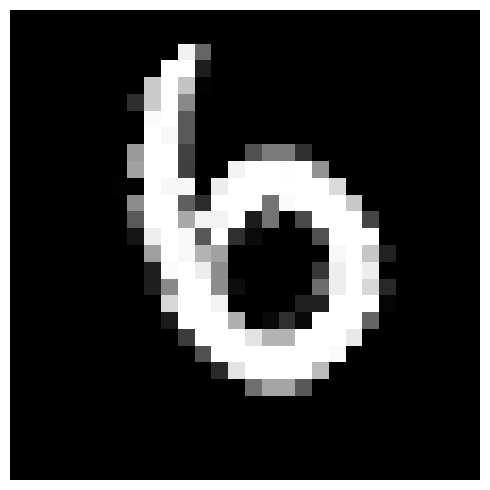}\\[-4pt]
        {\vspace{-2pt}\footnotesize$\scriptstyle \alpha_{\mathcal X}(a)(x)$}
      \end{minipage}
      \\[2pt]
      
      \begin{minipage}
      {\figwidth}\centering
      \vspace{1pt}
        \includegraphics[width=\figwidth]{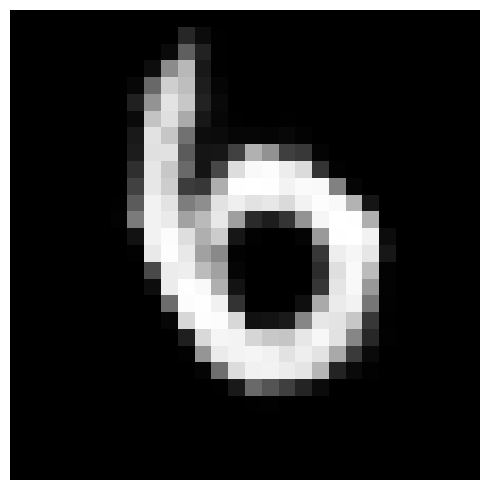}\\[-2pt]
        {\footnotesize$\scriptstyle D\bigl(\widehat\rho_{Z}(a)\,E(x)\bigr)$}
      \end{minipage}
      &
      \begin{minipage}{\figwidth}\centering
      \vspace{1pt}
        \includegraphics[width=\figwidth]{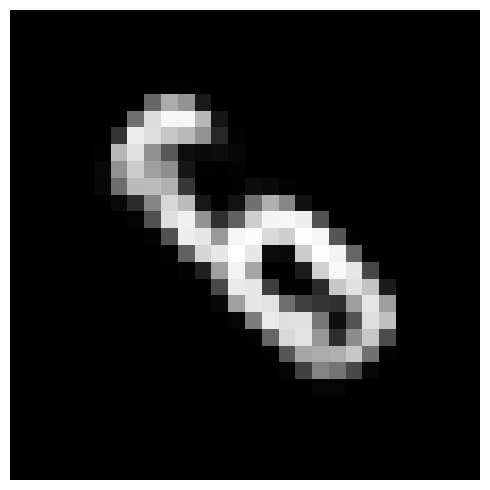}\\[-2pt]
        {\footnotesize$\scriptstyle D\bigl(\widehat\rho_{Z}(a)^2\,E(x)\bigr)$}
      \end{minipage}
    \end{tabular}

    \caption{Visualization of learned $E$, $D$ and $\widehat\rho_{Z}$ on a sample $x$ with $\alpha_{\mathcal X}$ swapping fonts. The algebraic loss correctly enforced $\widehat{\rho}_{Z}(a)^2=I_d$ while retaining equivariance.}
    \label{fig:visualisation}
  \end{minipage}
\end{wrapfigure}
For our first experiment we use the TMNIST~\citep{magre2022typographymnisttmnistmniststyleimage}dataset, of digits rendered in a variety of typefaces. We choose a subset of two typefaces only, producing 20 images, augmenting with 180° rotations and random scaling in $(0.8,1.2)$. For our group we choose $G=C_2$ presented as $\{1, a \, | \, a^2 = 1\}$. Since this is an autoencoder we have ${\mathcal X} = {\mathcal Y}$, and we choose $\alpha_{\mathcal X} = \alpha_{\mathcal Y}$, with the nontrivial element $\alpha_{\mathcal X}(a)=\alpha_{\mathcal Y}(a)$ acting to flip the choice of font, with rotation and scaling left invariant. For the algebra loss component \iveq we choose $\ALG_{C_2,d} = \MSE(\widehat{\rho}_{Z}(a)^2, \mathrm{I}_d)$ where $d=\dim({Z})=8$.

 Table~\ref{tab:tmnist} shows our findings, with each run giving one row of the table, and Figure~\ref{fig:visualisation} shows a visualization. Low values in the algebra and equivariance loss columns reveal high-quality representations $\widehat \rho_{Z}$, which are strongly equivariant with respect to the representations $\alpha_{\mathcal X}, \alpha_{\mathcal Y}$. By mapping the eigenvalues of $\widehat \rho_{Z}(a)$ to the nearest value in $\{-1, +1\}$, we can determine the corresponding irreducible representation. For the group $C_2$ the regular representation contains one copy of the -1 and +1 representations, and the learned $\widehat\rho_Z$ are close to a multiple of the regular representation. Furthermore, we report the number of linearly independent embedded orbits and, as expected, this corresponds to the number of copies of the regular representation found (Section~\ref{sec:theory}).

\paragraph{MLP Equivariant task, non-abelian group group, geometric action.}
\begin{wrapfigure}[16]{r}{0.5\textwidth} 
  \vspace{-10pt}
  \setlength{\intextsep}{50pt}   
  \centering
  \begin{minipage}{0.5\linewidth}
    \includegraphics[width=\linewidth]{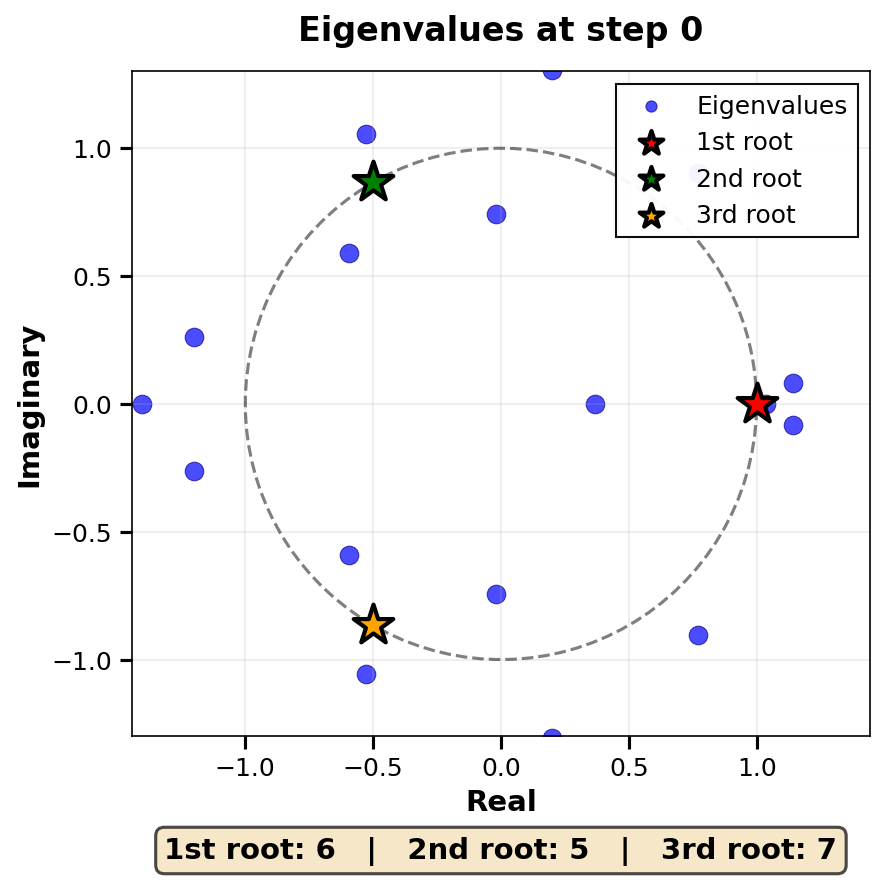}
  \end{minipage}\hfill
  \begin{minipage}{0.5\linewidth}
    \includegraphics[width=\linewidth]{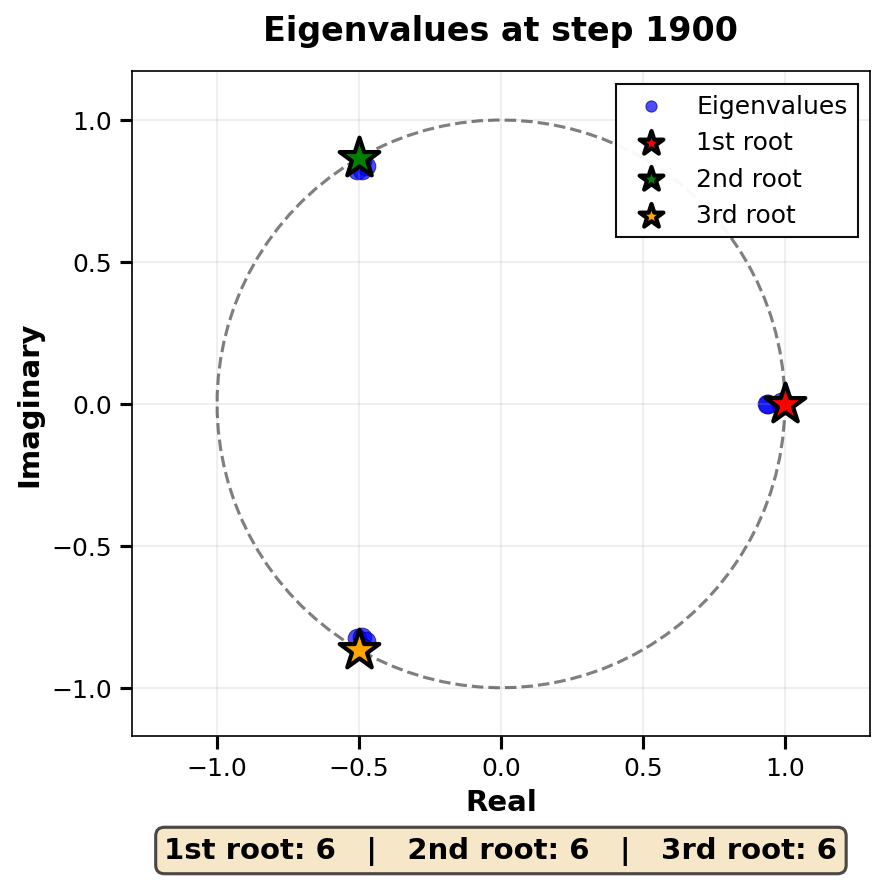}
  \end{minipage}

  \caption{Eigenvalues of $\widehat\rho_Z(r)$ over training. Counts near third roots of unity (bottom) show convergence to a uniform distribution despite uneven initialization.}
  \label{fig:d3-eigenvalues}
\end{wrapfigure}

For our second experiment we choose the MNIST~\citep{deng2012mnist} dataset, with augmentations given by arbitrary rotations. 
We choose the group $G = D_3$ of symmetries of an equilateral triangle with the generators $r,s$ (acting as 120-degree rotation and flip, respectively) and the following presentation: 
$\{e,r,r^2,r^3,s,rs\,|\,r^3=e, s^2=e, rsrs=e\}$.
We parameterize the linear maps $\widehat \rho_{Z}(r)$ and $\widehat\rho_{Z}(s)$ independently, and define the following algebra loss, where $d = \dim({Z}) = 18$, and where summands correspond to constraints in the presentation:
    $\ALG_{D_3,d}= \MSE(\widehat{\rho}_{Z}(r)^3, \mathrm{I}_d)
    +
    \MSE(\widehat{\rho}_{Z}(s)^2, \mathrm{I}_d) + \MSE(\widehat{\rho}_{Z}(r)\widehat{\rho}_{Z}(s)\widehat{\rho}_{Z}(r) \widehat{\rho}_{Z}(s)
    , \mathrm{I}_d)$.

For the nonabelian group $D_3$, we determined the learned representation's composition using orthogonality of characters (Section~\ref{sec:background}). The data in Table~\ref{tab:tmnist} confirms that the network learns a high-fidelity multiple of the regular representation, which contains the trivial, sign, and standard irreducible representations in the ratio 1:1:2. Consistent with the previous experiment, each linearly independent data orbit contributes one distinct copy of this representation. Furthermore, Figure~\ref{fig:d3-eigenvalues} illustrates the eigenvalues of the generator $\widehat \rho_{Z}(r)$ dynamically clustering around the third roots of unity during training, despite an uneven initialization.

\begin{table}[h]
\vspace{-5pt}
    \small
    \setlength{\tabcolsep}{1mm}
    \centering
    \caption{CIFAR classifier task \textcolor{black}{with analytic encoder, representations of $C_4$ learned on latent space. $Z$ is taken as the main feature layer before the final classification head.}\label{tab:cifar}}
    \vspace{-3pt}
    \begin{tabular}{cccccccc}
    \toprule
    &\multicolumn{4}{c}{Irreducible counts}
    \\
    \cmidrule{2-5}
        Run & $+1$ & $+i$ & $-1$ & $-i$ & Alg.\ loss & Eq.\ loss & Orbs.
        \\ \midrule
        1 & 4 & 4 & 4 & 4 & $1.5 \newtimes 10^{-4}$ & $1.8\newtimes 10^{-3}$ & 4 \\
        2 & 3 & 4 & 5 & 4 & $7.2 \newtimes 10^{-5}$ & $1.9\newtimes 10^{-3}$ & 3 \\
        3 & 3 & 5 & 3 & 5 & $9.4 \newtimes 10^{-5}$ & $1.6\newtimes 10^{-3}$ & 3 \\
        4 & 4 & 4 & 4 & 4 & $1.1 \newtimes 10^{-4}$ & $1.9\newtimes 10^{-3}$ & 4 \\
        5 & 4 & 4 & 4 & 4 & $8.4 \newtimes 10^{-5}$ & $1.9\newtimes 10^{-3}$ & 4
        \\ \bottomrule
    \end{tabular}
    \vspace{-10pt}
\end{table}%

\paragraph{CNN Invariant task, abelian group, geometric action}

This experiment uses the CIFAR10 dataset~\citep{Krizhevsky2009LearningML}.
We choose the abelian group $G = C_4$ of 90-degree rotations, with the algebraic loss function $\ALG_{C_4,d} = \MSE\left(\widehat{\rho}_{Z}(1)^4, \mathrm{I}_d\right)$, where $d = \dim({Z}) = 16$.
For $C_4$ the regular representation contains exactly one copy of the $+1$, $+i$, $-1$ and $-i$ representations, and Table~\ref{tab:cifar} shows that the network learns a representation close to a multiple of the regular representation.
Furthermore, each linearly independent embedded data orbit contributes a distinct copy of this representation.


\begin{tcolorbox}[highlightbox]
\textbf{Key empirical insight:} 
To achieve encoder equivariance in the presence of data augmentation, the network prefers to learn a multiple of the regular representation on the latent space.
\end{tcolorbox}

\section{Fixing the regular representation (\textbf{RQ3})}\label{sec:method}

Inspired by the {theoretical and empirical} results of Section~\ref{sec:optimal-rep}, \textbf{instead of learning a representation on the latent space, we now fix $\rho_{Z}$ to be a multiple of the regular representation of $G$} as an algebraic inductive bias. Specifically, we use $n$ copies where $n$ is the maximum number of representations allowed by $\text{dim }Z$. When $n|G| < \dim({Z})$, we pad by taking the direct sum with additional copies of the trivial representation, to ensure our representation on ${Z}$ has the correct dimension, yielding:
\begin{equation}\label{Eq:method}
    \rho_{Z} := n \cdot \rho_\mathrm{reg} + \mathrm{max}(\dim({Z}) - n|G|,0) \cdot \rho_\mathrm{triv}
\end{equation}
When the latent space is geometrically structured, e.g., as a product of features and channels, 
we choose an isomorphic form of the regular representation that preserves this structure (examples are the SMOKE and SHREC experiment in Section \ref{sec:experiments}). 
Denoting $(x_i, y_i) \in {\mathcal X} \times {\mathcal Y}$ an element of the training set, $g \in G$ a group element, and $L_\mathrm{task}(x_i, y_i)$ the task loss function, we train according to:\allowdisplaybreaks
\setlength{\jot}{3pt}
\begin{equation*}
\begin{alignedat}{2}
  &\tfrac12\,L_{\mathrm{task}}\bigl(D(E(x_i)),y_i\bigr)
  &&\quad\text{Task loss} \\ 
  &+ \tfrac12\,L_{\mathrm{task}}\bigl(D(E(\alpha_\mathcal{X}(g)x_i)),\,\alpha_\mathcal{Y}(g)y_i\bigr)
  &&\quad\text{Task loss shifted by $g$} \\
  &+ \lambda\,\mathrm{MSE}\bigl(E(\alpha_\mathcal{X}(g)x_i),\,\rho_{{Z}}(g)\,E(x_i)\bigr)
  &&\quad
    \begin{tabular}[t]{@{}l@{}}
      Equivariance loss from input to latent
    \end{tabular}
\end{alignedat}
\label{eq:training-loss}
\end{equation*}
When used in a training loop, we select $(x_i,y_i)$ and $g$ uniformly at random. 
Here $\lambda$ is a  hyperparameter expressing the strength of the equivariance loss. We provide a sensitivity analysis for $\lambda$ in Appendix \ref{Appendix:sensitivity}, which shows that model performance is robust across a range of values.
Our model has no additional learned parameters above baseline, since the representation $\rho_{Z}$ is now fixed. Our use of the $g$-shifted task loss means that our training dataset must be augmented by the action of $G$. This can be done either on-the-fly, or pre-computed to speed up training.







\subsection{Experiments}
\label{sec:experiments}

We benchmark our method across four distinct tasks against a comprehensive suite of state-of-the-art approximate equivariance methods, including SCNN, E2CNN, LIFT, RPP, RGroup, RSteer, PSCNN, NIso, NFT, and MC. For all comparisons, we adhere to the experimental setups described in the original papers. Our results demonstrate that our approach frequently outperforms these complex baselines while utilizing simpler architecture with fewer parameters and a comparable or lower computational budget. To isolate our contributions, we also include both augmented and unaugmented CNN baselines. We quantify the significance of our gains using Cohen's $d$-statistic~\citep[p59]{hermann2024analysis}, which consistently indicates very large performance improvements (see discussion in Appendix \ref{appendix:cohen-stat}). The equivariance loss is applied to the encoder output for autoencoding tasks and the penultimate layer for classification. Finally, to validate the optimality of the regular representation, we provide ablations replacing it with the trivial representation and the \textit{defining representation} (defined formally in Appendix~\ref{Appendix:group-actions}). Full technical details, including hyperparameter selection and a sensitivity analysis for the coupling strength $\lambda$, are provided in Appendices \ref{appendix:main-exp} and \ref{Appendix:sensitivity}.

\paragraph{Invariant classification task, DDMNIST.}
\label{sec:invariant_classification}
Following closely the procedure of~\citep{Veefkind2024} for each of the chosen symmetry groups $C_2$, $C_4$ and $D_4$, we randomly and independently transform two \mnist images according to the group. Results are shown in Table \ref{tab:ddmnist}.
Because the transformations are local and independent, we apply our method using the product group.
\textcolor{black}{We also provide a comparison with the defining and trivial representations as an ablation study.} While for the groups $C_2$ and $C_4$ the \textcolor{black}{two} representations are isomorphic, for $D_4$ they are not, with the regular representation being more performant; this provides further empirical evidence for the optimality of the regular representation. Except for SCNN, we re-trained and re-evaluated all models. Further discussion \textcolor{black}{and effect size analysis can be found in Appendix \ref{sec:ddmnist-appendix}. These statistics show a very large effect size for our model over the CNN baseline, and a large effect size for our model compared to the majority of results for other architectures.} 
\begin{table*}[h]\small
\setlength{\tabcolsep}{1mm} \caption{DDMNIST test accuracies. Mean over 3 runs; standard deviation in brackets. Parameter counts shown. Best result in each column is bold, second-best is underlined. For $C_2, C_4$ the defining representation is equivalent to the regular representation and so is omitted.}\label{tab:ddmnist}
\centering
\vspace{-5pt}
\begin{tabular}{lcc@{\hspace{6mm}}cc@{\hspace{6mm}}cc} \toprule {Model} & {$C_4\uparrow$} & {\#Params(M)$\,\downarrow$} & {$C_2\uparrow$} & {\#Params(M)$\,\downarrow$} & {$D_4\uparrow$} & {\#Params(M)$\,\downarrow$} \\\midrule CNN & 0.907 \sd{0.004} &\textbf{0.03}& \underline{0.938} \sd{0.006} &\textbf{0.03}& 0.800 \sd{0.001} &\textbf{0.03} \\ SCNN & 0.484 \sd{0.008} &\underline{0.12}& 0.474 \sd{0.003} &\textbf{0.03}& 0.431 \sd{0.010} &\underline{0.15} \\ Restriction & \underline{0.914} \sd{0.007} &\underline{0.12}& 0.890 \sd{0.007} &0.33& 0.837 \sd{0.020} &0.17 \\ RPP & 0.908 \sd{0.022} &0.79& 0.903 \sd{0.009} &0.08& 0.827 \sd{0.020} &1.73 \\ PSCNN & 0.909 \sd{0.007} &0.51& 0.871 \sd{0.016} &0.04& \underline{0.842} \sd{0.011} &1.23 \\ \midrule Trivial rep & 0.874 \sd{0.004} &\textbf{0.03}& 0.938 \sd{0.007} &\textbf{0.03}& 0.819 \sd{0.004} & \textbf{0.03}
\\  Defining rep & \qquad -- && \qquad -- && 0.838 \sd{0.010} &\textbf{0.03}
\\ Ours (regular) & \textbf{0.915} \sd{0.004} &\textbf{0.03}& \textbf{0.947} \sd{0.004} &\textbf{0.03}& \textbf{0.868} \sd{0.002} &\textbf{0.03} \\ \bottomrule \end{tabular}
\vspace{-10pt}
\end{table*}

\paragraph{Approximately equivariant classification Task, MedMNIST3D.}
\label{sec:medmnist}

We test our method on the Organ, Synapse and\ Nodule subsets of the \medmnist dataset, using the same setup as the original authors~\citep{Veefkind2024}. We apply the group $\mathrm{Sym}_{\mathrm{cube}}$ of orientation-preserving symmetries of the cube, which is isomorphic to the permutation group $S_4$. All results, except for ours and the augmented CNN, are imported from the original authors. Table \ref{tab:medmnist} shows \medmnist accuracies for different models and groups.
For Nodule and Synapse, our method is comparable or outperforms other architectures, while having  fewer parameters. The regular representation consistently outperforms the defining and trivial representations, providing further empirical evidence for its optimality.
For the Organ dataset, canonical orientation is a key feature, and so the symmetry action conflicts with the task. This may explain our method's underperformance in this task (shared by the augmented CNN baseline). \textcolor{black}{Further discussion can be found in Appendix~\ref{sec:medmnist-appendix}, which shows our method has very large positive effect sizes for Nodule and Synapse datasets.}

\begin{table}[h]
\small
\vspace{-5pt}
\setlength{\tabcolsep}{1mm} \caption{MedMNIST3D test accuracies. Mean over 3 runs; standard deviation in brackets. Parameter counts shown. Best result in each column is bold, second-best is underlined.\label{tab:medmnist}} \centering 
\vspace{-5pt}
\begin{tabular}{llc@{\hspace{4mm}}c@{\hspace{4mm}}c@{\hspace{4mm}}c@{\hspace{0mm}}c} \toprule Group & Model & Nodule$\,\uparrow$ & Synapse$\,\uparrow$ & Organ$\,\uparrow$ & \#Params(M)$\,\downarrow$ \\ \midrule N/A & CNN & 0.873 \sd{0.005} & 0.716 \sd{0.008} & 0.920 \sd{0.003} & \underline{00.19}\\ Aug & CNN & \underline{0.879} \sd{0.007} & 0.761 \sd{0.008} & 0.632 \sd{0.005} & \underline{00.19}\\ SO(3) & SCNN & 0.873 \sd{0.002} & 0.738 \sd{0.009} & 0.607 \sd{0.006}& \textbf{00.13} \\ SO(3) & RPP & 0.801 \sd{0.003} & 0.695 \sd{0.037} & \underline{0.936} \sd{0.002} & 18.30\\ SO(3) & PSCNN & 0.871 \sd{0.001} & \textbf{0.770} \sd{0.030} & 0.902 \sd{0.006} & 04.17\\ O(3) & SCNN & 0.868 \sd{0.009} & 0.743 \sd{0.004} & 0.902 \sd{0.006} & \underline{00.19}\\ O(3) & RPP & 0.810 \sd{0.013} & 0.722 \sd{0.023} & \textbf{0.940} \sd{0.006}& 29.30 \\ O(3) &PSCNN & 0.873 \sd{0.008} & \underline{0.769} \sd{0.005} & 0.905 \sd{0.004} & 03.51\\ \midrule $\mathrm{Sym}_{\mathrm{cube}}$ & Trivial rep & 0.867 \sd{0.001} & 0.743 \sd{0.002} & 0.571 \sd{0.002} & \underline{00.19}\\ $\mathrm{Sym}_{\mathrm{cube}}$ & Defining rep & 0.837 \sd{0.013} & 0.756 \sd{0.019} & 0.560 \sd{0.025} & \underline{00.19}\\ $\mathrm{Sym}_{\mathrm{cube}}$ & Ours (regular) & \textbf{0.887} \sd{0.005} & \textbf{0.770} \sd{0.002} & 0.642 \sd{0.056} & \underline{00.19} \\ \bottomrule \end{tabular}
\vspace{-5pt}
\end{table}

\paragraph{Approximately equivariant autoregression task.}
\label{sec:smoke}

We evaluate our method on the \smoke dataset, generated with PhiFlow~\citep{holl2020phiflow} by~\cite{wang2022approximately} (see Figure~\ref{fig:smokevis} for a visualisation). 
The task involves predicting future frames of a simulated smoke velocity field autoregressively. This task is only approximately equivariant to the symmetry group $C_4$ (90-degree rotations) due to the presence of non-equivariant buoyancy effects. Full details are provided in the appendix.
Table \ref{tab:smoke}(a) shows the test RMSE for each model on the metrics considered.
All reported figures are imported from the original authors~\citep{wang2022approximately}, except for ours, augmented CNN, and non-augmented CNN, for which we tune the learning rate.
Our method outperforms all models except for PSCNN, which has slightly better scores, with more than 12 times the number of parameters. While our method uses the augmented training set, we note from comparing the two CNN baselines that this gives little advantage for this task. \textcolor{black}{Further details can be found in Appendix \ref{sec:smoke-appendix}, showing very large positive effect sizes for all models except PSCNN.}

\paragraph{Equivariant autoencoding task, 3D shapes.}\label{sec:shrec}
Finally, we test our method on the conformally transformed SHREC '11 dataset~\citep{lianSHREC11TrackShape2011, mitchelMobiusConvolutionsSpherical2022}, following the pre-training and fine-tuning procedure of~\cite{mitchelNeuralIsometriesTaming2024}.
We apply our methodology with $O_h$ augmentations (octahedral symmetries) to pre-train a baseline autoencoder before fine-tuning the encoder for classification. As this is an autoencoding task, we symmetrize the equivariance loss to the decoder.
NIso's kernel adds 18k parameters above our model, which has the same parameter count as the baseline autoencoder (AE). Results are given in Table~\ref{tab:smoke}(b). Our approach achieves 90.45\% accuracy, outperforming the group-agnostic method NFT. Our method also surpasses NIso, a model capable of learning actions of infinite groups, even though our method uses only a finite subgroup. \textcolor{black}{Further details can be found in Appendix \ref{Appendix:shrec}, with effect sizes showing equivalence between our method and NIso, and very large positive effect size for the other models.}
\begin{table}[h]
\small
\setlength{\tabcolsep}{1mm}
    \centering
       \caption{Mean over 3 runs; standard deviation in brackets. Parameter counts shown. Best result in each column is bold, second-best is underlined.}\label{tab:smoke}
\vspace{-4pt}
\begin{tabular}{ccc}
    \begin{tabular}{l@{\hspace{3mm}}l@{\hspace{3mm}}l@{\hspace{3mm}}cc}
        \makebox[0pt][l]{\hspace{40pt}(a) Test RMSE for \smoke dataset.}
        \\[5pt]
        \toprule
        Group & Model  & Future$\,\downarrow$ & Domain$\,\downarrow$ & \#Params(M)$\,\downarrow$
        \\ \midrule
        N/A & CNN  & 0.81 \sd{0.01} & 0.63 \sd{0.00} & \textbf{0.25}
        \\
        Aug & CNN  & 0.83 \sd{0.03} & 0.67 \sd{0.06} & \textbf{0.25}
        \\
        N/A & MLP  & 1.38 \sd{0.06} & 1.34 \sd{0.03} & 8.33
        \\
        C4 & E2CNN  & 1.05 \sd{0.06} & 0.76 \sd{0.02} & \underline{0.62}
        \\
        C4 & RPP  & 0.96 \sd{0.10} & 0.82 \sd{0.01} & 4.36
        \\
        C4 & Lift  & 0.82 \sd{0.01} & 0.73 \sd{0.02} & 3.32
        \\
        C4 & RGroup  & 0.82 \sd{0.01} & 0.73 \sd{0.02} & 1.88
        \\
        C4 & RSteer  & 0.80 \sd{0.00} & 0.67 \sd{0.01} & 5.60
        \\
        C4 & PSCNN  & \textbf{0.77} \sd{0.01} & \textbf{0.57} \sd{0.00} & 3.12
\\ \midrule
        C4 & Ours  & \underline{0.78} \sd{0.01} & \underline{0.61} \sd{0.01} & \textbf{0.25} \\ \bottomrule
    \end{tabular}
&\quad\quad&
    \begin{tabular}{ll}
        \makebox[0pt][l]{\hspace{-0pt}\makebox[5.0cm][c]{(b) Test accuracy for SHREC '11 dataset.}}
        \\[5pt]
\toprule
Model  & Acc.$\,\uparrow$
\\ \midrule
        NIso \cite{mitchelNeuralIsometriesTaming2024} & \underline{90.26} \sd{1.27}\\
        NFT \cite{koyamaNeuralFourierTransform2024a} & 83.24 \sd{2.03}\\
        AE with aug & 69.36 \sd{2.81}\\
        MC \cite{mitchelMobiusConvolutionsSpherical2022} & 86.5 \\ \midrule
        Ours & \textbf{90.45} \sd{2.1}\\
        \bottomrule
    \end{tabular}
\end{tabular}
\vspace{-10pt}
\end{table}




\section{Conclusions}

\vspace{-3pt}
\textbf{Limitations and Future Work.}
\label{sec:limitations-future-work} 
Our method requires data augmentation, although this is typically inexpensive when the group action on the input space is easy to construct, and our ablations with an augmented baseline show that our model delivers benefits far beyond augmentation.
We would also like to explore how our model could enable augmentation directly in the latent space. Another limitation is that our theoretical analysis is restricted to finite groups; however, we show empirically that applying our method to a finite subgroup yields competitive results. Future work could investigate the suitability of such subgroups and extend the theoretical analysis to infinite groups.

\textbf{Conclusions.}
\label{sec:conclusions}
This work investigates an alternative path to building efficient equivariant models, focusing not on architectural design, but on the enforcement of a principled latent algebraic prior. We prove that for finite groups, this structure is the regular representation, which must appear almost surely in the latent space of any approximately equivariant encoder. By enforcing this structure via an auxiliary loss, our method achieves competitive or superior performance to SOTA models, while requiring in some cases significantly fewer parameters. Furthermore, we empirically show the optimality of the regular representation via ablations with the defining and trivial representations.  Ultimately, our work suggests that for tasks with inherent (approximate) symmetry, directly enforcing the correct latent algebraic structure can be a more effective and efficient path to equivariance than designing complex, highly-parameterized architectures.



\section*{Acknowledgments}
The authors would like to acknowledge Ioannis Markakis, Francesco Caso, Christopher Irwin and Lorenzo Sani for their insightful feedback.

\bibliography{iclr2026_conference}
\bibliographystyle{iclr2026_conference}

\appendix

\newpage
\etocdepthtag.toc{appendix}

\etocsettagdepth{main}{none}
\etocsettagdepth{appendix}{subsection}
\renewcommand{\contentsname}{Appendix}

\tableofcontents

\section{Code}\label{Appendix:code}
The code to run all the experiments in this paper is available at the following location:
\begin{itemize}
\item[]
https://github.com/rick-ali/parameter-free-approximate-equivariance
\end{itemize}
In the README file, we provide instructions to run the code and reproduce the results.

\section{Extended related work}\label{Appendix:extended-related-work}
A wide variety of methods have been developed to train neural networks to solve tasks in the presence of invariance, equivariance, or approximate equivariance. We give a brief summary here of those methods which are most relevant for our present work.

One of the most studied bodies of work derive from Convolutional Neural Networks (CNNs), which of course have strict translation invariance in their traditional form~\citep{LeCun1998, Shorten2019}. \cite{cohen2016steerablecnns} employ the framework of steerable functions~\citep{hel1998canonical} to construct a rotation-equivariant Steerable CNN\ architecture (\textbf{SCNN}), which strictly respects both translation and rotation equivariance; this was later generalised to develop a theory of general E(2)-equivariant steerable CNNs (\textbf{E2CNN}), which allow the degree of equivariance to be controlled by explicit choices of irreducible representation of the symmetry group~\citep{e2cnn}. Such a  network can avoid learning redundant rotated copies of the same filters. A similar method is that of Mobius Convolutions (\textbf{MC})~\citep{mitchelMobiusConvolutionsSpherical2022}. \cite{wang2022approximately} use steerable filters to obtain convolution layers with approximate translation symmetry and without rotation symmetry (\textbf{RSteer}), and with approximate translation and rotation symmetry (\textbf{RGroup}). These authors  relax the strict weight tying of E2CNNs, replacing single kernels with weighted linear combinations of a kernel family, with coefficients that are not required to be rotation- or translation-invariant. A third approach named Probabilistic Steerable CNNs (\textbf{PSCNN}) was proposed recently by~\cite{Veefkind2024}, which allows SCNNs to determine the optimal equivariance strength at each layer as a learnable parameter. 
\textcolor{black}{While equivariant architectures may allow reduced parameter counts due to weight-tying, in practice many of these architectures require considerable additional parameter counts to achieve competitive performance (see parameter counts in Section~\ref{sec:experiments}).}

We also discuss a  family of approaches which are not based around the CNN architecture. Residual Pathway Priors \textbf{(RPP)} \citep{NEURIPS2021_fc394e99}, is a model where each layer is doubled, yielding a first layer with strong inductive biases,  and a second layer which is less constrained, with final  output is obtained as the sum of these layers. Another architecture is Lift Expansion \textbf{(LIFT)}, which factorizes the input space into equivariant and non-equivariant subspaces, and applies different architectures to each~\citep{wang2021equivariant}. 

\textcolor{black}{A number of previous studies have considered group representations on the latent space, sometimes governed via an equivariance term in the loss function. An early approach by~\citep{cohen2015transformation} shows how geometrical transformations can be encoded on the latent space via SO(3) representations on the latent space, while~\citep{Worrall_2017} demonstrate disentanglement phenomena with similar methods.}
\cite{dupont2020equivariantneuralrendering} propose a parameter-free method to learn equivariant neural implicit representations for view synthesis; while similar to our method in some respects, such as fixing the latent representation, their work strongly leverages the defining representation of the infinite group $O(3)$, is limited to latent spaces with the same geometrical structure as the input space, and does not apply to arbitrary latent encodings. \cite{jin2024learning} present a similar method  which learns non-linear group actions on the latent space using additional learnable parameters, augmented by an optional attention mechanism. In Neural Isometries (\textbf{NIso})~\citep{mitchelNeuralIsometriesTaming2024}, the authors propose to learn an action on the latent space via its eigenbasis; in contrast, in our model the group acts linearly on the latent space with a fixed representation, and with no additional parameters needed. \textcolor{black}{Recent work of~\citep{Bokman_2024} considers learned latent representations for a fixed group to solve certain geometrical tasks.} Other approaches that do not require the symmetry group to be known beforehand include Neural Fourier Transforms (\textbf{NFT})~\citep{koyamaNeuralFourierTransform2024a}, which seeks to learn a suitable latent space transformation, and other work ~\citep{shakerinavaStructuringRepresentationsUsing2022, winterUnsupervisedLearningGroup2024}.

While our work builds on these approaches, our contribution is distinct: by assuming a known symmetry, we leverage representation theory to identify the regular representation as a theoretically-motivated latent structure, which enables our simple pipeline without additional learnable parameters. \textcolor{black}{Although our approach requires fixing a group structure, our experimental results show that this approach can often achieve superior performance compared to models without this constraint, including models specifically adapted for continuous symmetries.}

\section{Notation}\label{Appendix:notation}
Here we provide a comprehensive list of symbols and notational conventions used throughout the paper.

\subsection*{General mathematical objects}
\begin{description}
    \item[$G$] A finite group.
    \item[$g, h$] Elements of the group $G$, e.g., $g \in G$.
    \item[$\mathbb{K}$] The base field, assumed to be either the real numbers $\mathbb{R}$ or the complex numbers $\mathbb{C}$.
    \item[$\mathcal{S}, \mathcal{A}$] General sets, denoted by calligraphic letters.
    \item[$V, W$] General vector spaces, denoted by uppercase Roman letters.
    \item[$v, w$] Elements (vectors) of a vector space, e.g., $v \in V$.
    \item[$f|_\mathcal S$] If $f:\mathcal A\to \mathcal B$ is a function and $\mathcal S\subseteq\mathcal A$, $f|_\mathcal S$ denotes the restriction of the function to $\mathcal S$.
\end{description}

\subsection*{Group theory}
\begin{description}
    \item[$\alpha$] A group action on a set. The action of $g \in G$ on an element $s \in \mathcal{S}$ is written as $\alpha(g, s)$ 
    \item[$\rho_V$] A group representation on the vector space $V$, which is a linear group action on $V$.
    \item[$\rho_V(g)$] The invertible linear map associated with the group element $g \in G$. The action of $g$ on a vector $v \in V$ is written as $\rho_V(g)(v)$.
\end{description}

\subsection*{Machine learning context}
\begin{description}
    \item[$\mathcal{X}$] The input set.
    \item[$x$] A single input data point, $x \in \mathcal{X}$.
    \item[$\mathcal{Y}$] The output or label set.
    \item[$y$] A single output or label, $y \in \mathcal{Y}$.
    \item[$Z$] The latent space, viewed as a vector space (e.g., $Z = \mathbb{R}^d$).
    \item[$z$] A latent vector, $z \in Z$.
    \item[$E$] An encoder network.
    \item[$D$] A decoder network.
    \item[$\widehat{\rho}_Z$] A \textit{learnable} representation on the latent space $Z$.
\end{description}

\section{Group actions and representations}\label{Appendix:group-actions}

\paragraph{Groups.} A group $G$ is a set equipped with an associative and unital binary operation, such that every element has a unique inverse. 
Important families of groups include the following. The dihedral group $D_n$ is the group of symmetries of the regular polygon with $n$ sides, which we use in this paper for $n \geq 3$. The cyclic group $C_n$ is the groups of integers $\{0, \ldots, n-1\}$ with addition modulo $n$. The permutation group $S_n$ is the group of permutations of an $n$-element set. We may define groups by presentations, which give generators and relations for the product; for example, the group $C_2$ can be defined by the presentation $\{1, a \,|\, a^2 = 1\}$. For any two groups $G,H$, we write $G \times H$ for the product group, whose elements are ordered pairs of elements of $G$ and $H$ respectively.

\paragraph{Group representations.} A \textit{representation} $\rho$ of a finite group $G$ on a vector space $V$ is a choice of linear maps $\rho(g):V \to V$ for all elements $g \in G$, with the property that $\rho(e) = \mathrm{id}_V$ for the identity element $e \in G$, and such that $\rho(g)\rho(g') = \rho(gg')$ for all pairs of elements $g,g' \in G$. We define the \textit{dimension} of $\rho$ to be $\text{dim}(V)$, the dimension of the vector space $V$. There is a notion of equivalence of representations: given representations $\rho$ on $V$, and $\rho'$ on $V'$, they are \textit{isomorphic} when there is an invertible linear map $L: V \to V'$ such that $L\rho(g) = \rho'(g)L$ for all $g \in G$. Given a subgroup $G \subseteq G'$, a representation of $G'$ yields a \textit{restricted representation} on $G$ in an obvious way.

\paragraph{Defining representations.} The concept of a defining representation is relevant for our ablation studies. While the term is context-dependent, it typically refers to a group's most natural or defining low-dimensional representation. For the permutation group $S_n$ this is the linearisation of its permutation action on the $n$-element set; that is, the $n$-dimensional representation given by its action on $\mathbb{K}^n$ by permuting the basis vectors. For the dihedral group $D_n$ ($n \ge 3$), the defining representation is the linearisation of its action on the $n$-element set of vertices. For the group $\mathrm{Sym}_\mathrm{cube}$ of orientation-preserving symmetries of the cube, the defining representation is the linearisation of its action on the 8-element set of vertices of the cube. We select these defining representations as a baseline as they provide a rich, geometrically intuitive alternative to the more abstract regular representation.

\paragraph{Group actions.} A group may also have an \textit{action} $\lambda$ on a set $\mathcal S$, a choice of functions $\lambda(g): \mathcal S \to \mathcal S$ for all elements $g \in G$, such that $\lambda(e) = \mathrm{id}_S$ and $\lambda(g)\lambda(g') = \lambda(gg')$. Such an action yields a representation of $G$ on $\K[\mathcal S]$ by linearisation, the \textit{free \K-vector space} generated by $S$.

Some simple examples of representations include the \textit{zero representation} on the zero-dimensional vector space, and the \textit{trivial representation} $\rho_\mathrm{triv}$ on the 1-dimensional vector space $\K$, where $\rho_\mathrm{triv}(g)=\mathrm{id}_\K$ for all $g \in G$. 



\section{Insight into the algebra loss}\label{Appendix:algebra-loss}
To give further insight into component \iveq,  suppose our goal is to learn a representation $\widehat {\rho}_{Z}$ of the group $C_2$, which has group presentation $\{1, a \,|\, a^2 = 1\}$. Then $\widehat {\rho}_{Z}$ should satisfy $\widehat {\rho}_{Z}(1)=\mathrm{id}$ and $\widehat\rho_Z(a^2)=(\rho_Z(a))^2=\mathrm{id}$. To achieve this, we fix the parameter $\widehat {\rho}_{Z}(1)=\mathrm{id}$, and choose $\ALG_{C_2,d}$ and $\REG_{C_2,d}$ as follows, where $d=\mathrm{dim}({Z})$, the matrix  $\mathrm{I}_d$ is the identity of size $d \times d$:
\[
\begin{aligned}
  \ALG_{C_2,d} &= \MSE\bigl(\,\widehat{\rho}_{Z}(a)^2,\;I_d\bigr)\\
  \REG_{C_2,d} &= \MSE\bigl(\,\widehat{\rho}_{Z}(a),\;\widehat{\rho}_{Z}(a)^{-1}\bigr).
\end{aligned}
\]
We note that when $\ALG_{C_2,d}$ equals zero then $\widehat \rho_{Z}(a)^2 = \mathrm{I}_d$, and hence  $\REG_{C_2,d}$ will equal zero. In this sense, the regularisation term is algebraically redundant, but is found to improve training.

\section{Proofs}\label{Appendix:proofs}
\color{black}
{We first introduce some basic definitions}

\begin{definition}
    \textcolor{black}{Let $V$ be a vector space, and $W, W'\subseteq V$ be subspaces of $V$. $W$ and $W'$ are \emph{linearly independent} if $W\cap W'=0$.}
\end{definition}

\begin{definition}
\label{def:linind}
    \textcolor{black}{Let $G$ act on a set $\mathcal X$ via the group action $\alpha_X$. The \emph{orbit of $x\in\mathcal X$} is the set $\mathcal O_x=\{\alpha_{\mathcal X}(g)(x)\,|\,g\in G\}$. Given a vector space $Z$ and a function $E:\mathcal X\to Z$, we call the set $E(\mathcal O_x)=\{E(\alpha_{\mathcal X}(g)(x)\,|\,g\in G\}$ the \emph{embedded orbit of $x$ along $E$}. Two embedded orbits $E(\mathcal O_x), E(\mathcal O_{x'})$ are \emph{linearly independent} if their spans are linearly independent, that is if $\mathrm{Span}(E_\theta(\mathcal O_x)) \cap \mathrm{Span}(E_\theta(\mathcal O_{x'})) = \{0\}$.}
\end{definition}
\color{black}

\begin{definition}
\label{def:approx-equiv}
    Let $G$ be a group acting on the sets $\mathcal A$ and $\mathcal B\subseteq\mathbb R^d$ with the actions $\alpha_{\mathcal A}$ and $\alpha_{\mathcal B}$, respectively. Let $E:\mathcal A\to \mathcal B$ be a function of sets. We say that $E$ is 
    \emph{$(\alpha_{\mathcal A},\alpha_{\mathcal B}, \varepsilon)$-equivariant} if 
    \begin{equation*}
        \sup_{g\in G\,\,x\in \mathcal X} ||E(\alpha_{\mathcal A}(g)(x))-\alpha_{\mathcal B}(g)(E(x))||\leq\varepsilon
    \end{equation*}
    This quantity measures how far $E$ is from being equivariant with respect to the actions $\alpha_{\mathcal A}$ and $\alpha_{\mathcal B}$. When clear from the context, we will drop the dependency on $\alpha_{\mathcal A}$. 
\end{definition}

\begin{definition}
\label{def:operator-norm}
    Let $V,W$ be vector spaces with norms $||\cdot||_V$ and $||\cdot||_W$ respectively, and let $A:V\to W$ be a linear map between them. The \emph{operator norm} is defined by
    \begin{equation*}
        ||A||_{\text{op}} = \sup_{x\neq 0}\frac{||Ax||_W}{||x||_V}=\max_{||x||_V=1}||Ax||_W
    \end{equation*}
    Then, given two linear maps $A, B$, the quantity $||A-B||_{\text{}op}$ gives a distance of functions.
\end{definition}

\color{black}
The proof of Lemma \ref{lemma:zero-measure} adapts the argument in \cite{nikolaou2025languagemodelsinjectiveinvertible}, which uses measure-theoretic properties of analytic functions to demonstrate that transformers are almost everywhere injective. Although our focus here is not on transformers, most of their results require only real analyticity, and thus can be easily adapted to our case. Intuitively, a measure $\mu$ on a set $X$ quantifies the `size' or `volume' of subsets within $X$ (see e.g., \cite{fremlin2000measure} for a foundational treatment). In the context of $\mathbb{R}^p$, the Lebesgue measure $\lambda$ corresponds to the standard notion of Euclidean volume (e.g., it assigns the unit hypercube a measure of $1$).

\paragraph{Notation.} If $f:X\to X$ is a function, we will write $f^{\circ T}$ to indicate the consecutive application of $f$ for $T$ times. If $\Theta$ is a set equipped with a measure $\mu$, we will write $\theta\sim\Theta$ to indicate that $\theta$ is a random draw of an element of $\Theta$ according to the measure $\mu$.

\textcolor{black}{The following are well-known mathematical results and definitions.}
\begin{proposition}\label{prop:zero-sets-analytic}
    \textcolor{black}{Let $U\subseteq \mathbb R^m$ be open and connected, and let $f:U\to \mathbb R^n$ be an analytic function. If $f$ is not identically zero, then its zero set $Z(f):=\{x\in U\,|\,f(x)=0\}=f^{-1}(0)$ has Lebesgue measure zero in $\mathbb R^m$, i.e.~$\lambda(Z(f))=0$.}
\end{proposition}

\begin{definition}
    {Let $\mu$, $\nu$ be Borel measures on $\mathbb R^p$. We say that $\mu$ is \textit{absolutely continuous} with respect to $\nu$, written $\mu\ll\nu$, if for every Borel set $U$ we have}
    \begin{equation}
        \nu(U)=0\implies \mu(U)=0
    \end{equation}
\end{definition}

Since the Lebesgue measure $\lambda$ is the standard notion of Euclidean volume, then we can intuitively understand that $\mu \ll \lambda$ just when the measure $\mu$ assigns zero measure to every set with zero Euclidean volume. In this way, measures $\mu$ with $\mu \ll \lambda$ may behave in ways that accord with our intuition. Any standard sampling measure $\mu$ used to generate initial parameters for a neural network (e.g. from the normal or uniform distributions) will be likely to satisfy this property.

\textcolor{black}{The following critical lemma underlies our theoretical results, and we can explain it intuitively as follows. If we have some set of parameters $\mathcal W \subseteq \Theta$ for our neural network which is measure zero, then of course if we initialise the network with some parameters $\theta$ at random, the probability that $\theta \in \mathcal W$ is zero. But we then ask, if we update the parameters by gradient descent for finitely many steps, what is the probability that the optimised parameters are within the set $\mathcal W$?}

\begin{lemma}
\label{lemma:zero-measure}
    \textcolor{black}{Let $E_\theta$ be a parametrized function that is analytic for all parameters $\theta\in\Theta\subseteq\mathbb R^p$. Assume that the parameters are randomly initialized according to a distribution $\mu$ that is absolutely continuous with respect to the Lebesgue measure on $\mathbb R^p$, i.e.~$\theta_0\sim\mu$ with $\mu\ll\lambda$. Furthermore, assume an analytic loss function $\mathcal L$, and that the parameters are updated via gradient descent, i.e.~$\theta_{t+1}:=\Phi(\theta_t):=\theta_t-\eta\nabla\mathcal L(\theta_t)$, with $\eta\in(0,1)$. }

    \textcolor{black}{Let $\mathcal W\subseteq\Theta$ with $\lambda(\mathcal W)=0$. Then, for all $T\in\mathbb N$, $\mu(\{\theta_0\,|\,\Phi^{\circ T}(\theta_0)\in\mathcal W\})=0$.}
\end{lemma}
\begin{proof}
    \textcolor{black}{The proof uses standard analytic and measure-theoretic tools, and is an adaptation of the argument in the proof of \cite[Theorem C.1]{nikolaou2025languagemodelsinjectiveinvertible}.
    Let $\mathcal W\subseteq \Theta$ with $\lambda(\mathcal W)=0$. First, we apply \cite[Lemma C.6]{nikolaou2025languagemodelsinjectiveinvertible},  that $\lambda(\Phi^{-1}(\mathcal W))=0$. (Note that in this paper, Lemma~C.6 assumes the context of a transformer; however the proof only uses analyticity of the components, and thus the result holds in our case).
    Then, because $\mu\ll\lambda$, it follows that $\mu(\{\theta_0\,|\,\Phi(\theta_0)\in\mathcal W\})=\mu(\Phi^{-1}(W))=0$. By applying the same argument $T$ times, we find $\mu(\{\theta_0\,|\,\Phi^{\circ T}(\theta_0)\in\mathcal W\})=0$.}
\end{proof}

\color{black}
\color{black}
\textbf{Theorem \ref{thm:reg-rep}.}
    Let $G$ be a finite group acting on a set $\mathcal A$ with action $\alpha_{\mathcal A}$, and on a vector space $Z$ with a representation $\rho_Z$, with $\dim(Z) \geq |G|$.
    Suppose that the group acts freely and transitively on some subset $\mathcal S \subseteq \mathcal A$. Let $E_\theta$ be a $(\rho_Z,\varepsilon)$-equivariant encoder and $\sigma_{\text{min}}$ the smallest singular value of $E(\mathcal S)$. If $\sigma_{\min} > 0$, there exists a representation $(V,\tilde\rho)$ with $V\subseteq Z$ and $\tilde\rho\cong\rho_{\text{reg}}$ such that for all $g\in G$:
    \begin{center}
    $||\rho_{Z}(g)|_V -\tilde\rho(g)||_{\text{op}}\leq  \frac{\varepsilon}{\sigma_{\text{min}}}\sqrt{|G|}$\ \ 
\end{center}
\begin{proof}
Let $\mathbb{R}[\mathcal S]$ denote the vector space of all formal linear combinations of $\mathcal S$ with coefficients in $\mathbb{R}$. Because $\alpha_{\mathcal A}|_\mathcal S$ is free and transitive it must be equivalent to the action of $G$ on itself, and hence its linearization $\mathbb{R}[\mathcal S]$ carries the structure of the regular representation. We write this representation explicitly as  $\rho_{\mathbb{R}[\mathcal S]}(g)(\sum_s a_s s):=\sum_s a_s \alpha_\mathcal A(g)(s)$. 

Define the linearization of $E$ as the linear map $\tilde{E}: \mathbb{R}[\mathcal{S}] \to Z$ by mapping basis elements $e_s \mapsto E(s)$. The matrix representation of $\tilde{E}$ is exactly $M^\mathcal{S}=E(\mathcal S)$, the matrix whose columns are given by $\{E(s)\}_{s\in\mathcal S}$. Since $\sigma:=\sigma_{\min}(M^\mathcal{S}) > 0$ by assumption, $\tilde{E}$ is a linear isomorphism onto its image $V = \text{Im}(\tilde{E})$.

We define the action $\tilde{\rho}$ on $V$ by pushing forward the regular representation through $\tilde{E}$:
$$\tilde{\rho}(g) := \tilde{E} \circ \rho_{\text{reg}}(g) \circ \tilde{E}^{-1}$$
By construction, $(V, \tilde{\rho}) \cong (\mathbb{R}[\mathcal{S}], \rho_{reg})$, so $V$ carries the exact regular representation structure.

Let $\rho_V(g):=\rho_{Z}(g)|_V$ be the restriction of the linear map $\rho_Z$ to $V\subseteq Z$. We now bound the difference between the actions $\tilde\rho$ and $\rho_V$. For any vector $v \in V$ with $\|v\|=1$, we can write $v = \tilde{E}u$ for a unique $u \in \mathbb{R}[\mathcal{S}]$. Note that $\|u\| \le \frac{1}{\sigma}\|v\| = \frac{1}{\sigma}$ by definition of $\sigma$.

We compare the actions:
\begin{align*}
    \|\rho_V(g)v - \tilde{\rho}(g)v\| &= \|\rho_V(g)\tilde{E}u - \tilde{E}\rho_{reg}(g)u\| \\
    &= \| (\rho_V(g)\tilde{E} - \tilde{E}\rho_{reg}(g)) u \|
\end{align*}
The term in the parenthesis is the `equivariance error' of the linear map $\tilde{E}$. For a basis vector $e_s$:
$$\| \rho_V(g)\tilde{E}e_s - \tilde{E}\rho_{reg}(g)e_s \| = \| \rho_V(g)E(s) - E(g \cdot s) \| \le \varepsilon$$
where the last inequality is given by assumption. For a general vector $u = \sum a_s e_s$, we have:
$$\| (\rho_V(g)\tilde{E} - \tilde{E}\rho_{reg}(g)) u \| \le \sum |a_s| \varepsilon = \varepsilon \|u\|_1$$
Using the inequality $\|u\|_1 \le \sqrt{|G|} \|u\|$, we get:
$$\| (\rho_V(g)\tilde{E} - \tilde{E}\rho_{reg}(g)) u \| \le \varepsilon \sqrt{|G|} \|u\| \le \varepsilon \sqrt{|G|} \frac{1}{\sigma}$$
for all $g\in G$. Thus, the operator norm difference is bounded by $\frac{\varepsilon}{\sigma} \sqrt{|G|}$.
\end{proof}

We now combine Lemma~\ref{lemma:zero-measure} and Theorem~\ref{thm:reg-rep} to obtain guarantees on the existence of regular representations in the latent space.

\textbf{Theorem \ref{thm:analytic}.}
Let $G$ be a finite group acting on a set $\mathcal A$ and on a vector space $Z$ with representation $\rho_Z$, where $\dim(Z) \geq |G|$. Let $\mathcal S \subseteq \mathcal A$ be an orbit on which $G$ acts freely.
Consider an encoder $E_\theta: \mathcal A \to Z$ parameterized by $\theta \in \Theta \subseteq \mathbb{R}^p$. Suppose $E_\theta$ is analytic with respect to $\theta$ and that the parameters are initialized $\theta_0 \sim \mu$ (with $\mu$ absolutely continuous w.r.t. Lebesgue measure) and updated via gradient descent on an analytic loss.

Then, exactly one of the following holds:
\begin{enumerate}[label=(\roman*), leftmargin=*, nosep]
    \item The encoder orbit $E_\theta(\mathcal S)$ is rank-deficient ($\text{rank} < |G|$) for all possible parameter values $\theta \in \Theta$.
    \item With probability 1, the latent subspace $V = \text{span}(E_\theta(\mathcal S))$ is isomorphic to the regular representation. Specifically, the orbit forms a basis for $V$, and the induced action $\tilde{\rho}$ on $V$ satisfies the bound from Theorem \ref{thm:reg-rep}:
    $\|\rho_Z(g)|_V - \tilde{\rho}(g)\|_{\text{op}} \leq \frac{\varepsilon}{\sigma_{\min}} \sqrt{|G|}$
\end{enumerate}

\begin{proof}
We analyze the rank of the matrix $M_\theta^{\mathcal S} = E_\theta(\mathcal S) \in \mathbb{R}^{\dim(Z) \times |G|}$ formed by the encoder outputs. The regular representation is instantiated if and only if these vectors are linearly independent, i.e., $\text{rank}(M_\theta^{\mathcal S}) = |G|$.

Let $d_I(\theta)$ denote the determinants of all possible $|G| \times |G|$ minors of $M_\theta^{\mathcal S}$. The condition of rank deficiency corresponds to the zero set $\mathcal{W} = \{ \theta \in \Theta \mid \forall I, d_I(\theta) = 0 \}$. Since the encoder $E_\theta$ is analytic, each determinant $d_I(\theta)$ is an analytic function of $\theta$.

By the properties of analytic functions, the common zero set $\mathcal{W}$ is either the entire domain $\Theta$ or a set of Lebesgue measure zero.
\begin{enumerate}[label=(\roman*), leftmargin=*, nosep]
    \item If $\mathcal{W} = \Theta$, the rank is strictly less than $|G|$ everywhere. The vectors are always linearly dependent, preventing the formation of the regular representation.
    \item If $\mathcal{W} \neq \Theta$, then $\mathcal{W}$ has measure zero. Since the initialization $\mu$ is absolutely continuous and the gradient update map is analytic (preserving null sets), the probability of the trajectory entering or remaining in $\mathcal{W}$ is 0. Thus, with probability 1, $\text{rank}(M_\theta^{\mathcal S}) = |G|$ and $V$ is isomorphic to the regular representation.
\end{enumerate}
In case (ii), the full rank implies $\sigma_{\min}(M_\theta^{\mathcal S}) > 0$. By Theorem \ref{thm:reg-rep}, this guarantees the existence of the subspace isomorphism and the validity of the stability bound.
\end{proof}
\color{black}

\subsection{\textcolor{black}{The analyticity condition}}\label{appendix:analyticity-condition}
\textcolor{black}{We remark that, as observed by \cite{nikolaou2025languagemodelsinjectiveinvertible}, most standard modules used in neural network, such as linear layers, layer norm, skip connections, convolutions, attention, and others are analytic. The same holds for many commonly used activation functions, such as $\text{tanh, sigmoid, softplus, softmax, SiLU, GELU, SwiGLU}$. Therefore, the analytic condition does not heavily restrict our analysis. For example, \cite{nikolaou2025languagemodelsinjectiveinvertible} highlight that decoder-only transformers are analytic. However, others activation functions are only piece-wise analytic, e.g.~$\text{ReLU, LeakyReLU, ELU}$. For this reason, we repeat the TMNIST and MNIST experiments from Section \ref{sec:empirical-exploration} with non-analytic encoders to empirically test whether our conclusions hold for this class of networks. We find this to be the case, and we discuss it in Appendix \ref{appendix:piecewise-analytic-exp}.
}

\section{Representational persistance}\label{Appendix:probes}
To evaluate the tension between the invariance objective and the theoretical persistence of the regular representation, we analyze the latent space of an encoder trained with an explicit invariance penalty: $\mathcal{L}_{\text{inv}} = \|E(x) - E(gx)\|^2$. We take $Z$ to be the layer before the classification head, and set $d=\dim(Z)=64$. For a given input $x$, we denote the latent embedding of its $g$-th transformation as $z_g = E(\alpha_{\mathcal X}(g)(x))$ and the orbit mean as $\bar{z} = \frac{1}{|G|} \sum_{g \in G} z_g$. We use these to define the residual vectors $r_g = z_g - \bar{z}$, which represent the portion of the signal that the encoder has failed to make invariant. We investigate whether the representation undergoes a structural transition to the trivial (invariant) representation. To quantify this, we use linear probes, a canonical framework in the interpretability literature~\citep{alain2018understandingintermediatelayersusing,hewitt-manning-2019-structural,belinkov2021probingclassifierspromisesshortcomings} used to assess whether specific latent properties are linearly decodable from the representation.

\textbf{Local linear probes.} We verify the existence of the regular representation by training a Local Linear Probe on each orbit. Given an orbit $\mathcal{O}_x = \{z_0, z_1, z_2, z_3\}$ in $\mathbb{R}^{64}$, we attempt to learn a linear mapping $W \in \mathbb{R}^{4 \times 64}$ to the canonical basis $\{e_1, e_2, e_3, e_4\} \subset \mathbb{R}^4$. The probe is given via the normal equation:$$W = Y X^T (X X^T + \lambda I)^{-1}$$ where $X$ is the matrix of orbit features and $Y$ is the identity matrix. We find that for all orbits, the reconstruction accuracy is $100\%$, and the weights $W$ remain well-conditioned. This empirically confirms that the residual vectors $\{z_g - \bar{z}\}$ are linearly independent, satisfying the full-rank condition $\sigma_{\min} > 0$ required by Theorem \ref{thm:analytic}.

\begin{figure}[t]
    \centering
    \begin{subfigure}{0.48\textwidth}
        \centering
        \includegraphics[width=\linewidth]{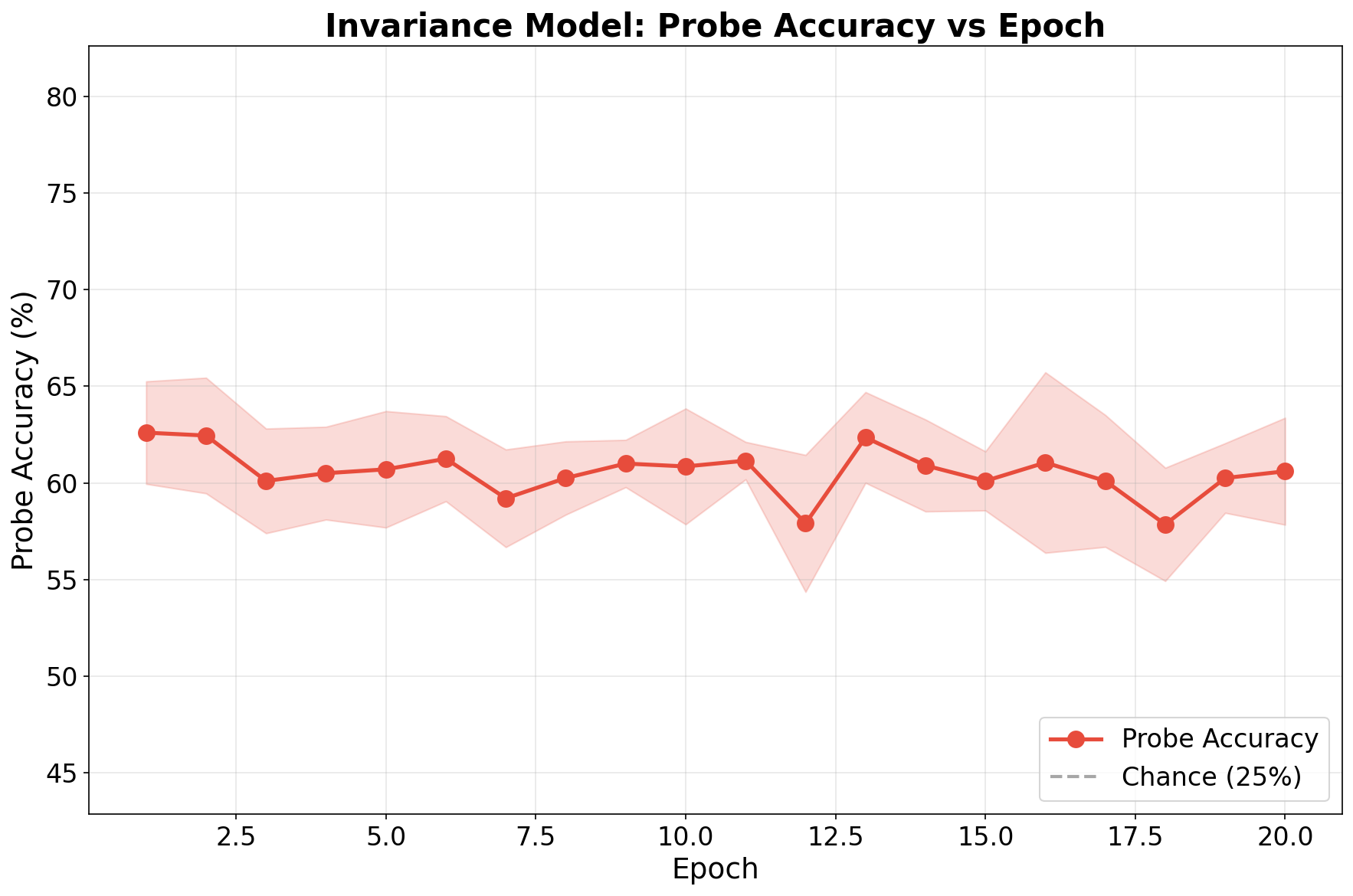}
        \caption*{}
        \label{fig:left}
    \end{subfigure}
    \hfill
    \begin{subfigure}{0.48\textwidth}
        \centering
        \includegraphics[width=\linewidth]{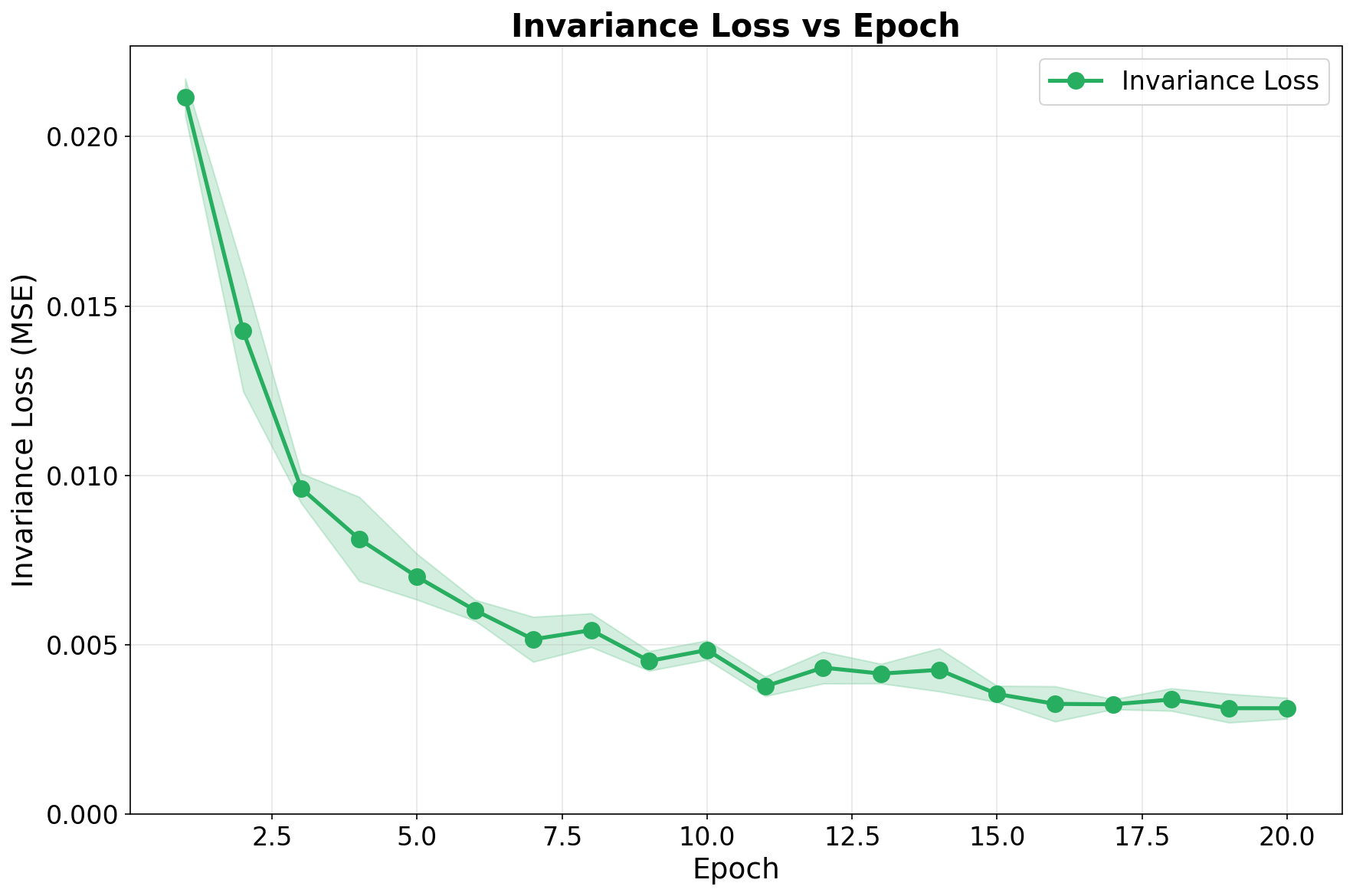}
        \caption*{}
        \label{fig:right}
    \end{subfigure}
    \vspace{-17pt}
    \caption{(Left) Global probe accuracy and (Right) invariance loss throughout the training of an invariant MLP. Despite the low invariance loss, the global probe maintains a significant accuracy (60.6\%$\pm$2.8\%), confirming that the regular representation persists in a decodable state rather than being structurally erased. Averages over 5 runs.}
    \label{fig:probe-accuracy}
\end{figure}

\textbf{Global Linear Probes.} To assess the consistency of this representation across the manifold, we train a Global Linear Probe. Unlike the local probe, which is fit to a single orbit, the global probe uses a single weight matrix $W_{\text{global}}$ trained on the entire training set to predict group elements from embeddings. Specifically, they are trained to predict $g$ from $r_g$ at each training epoch. On the held-out set, the global probe achieves 60.6\%$\pm$2.8\% accuracy (vs. 25\% chance), as shown in Figure \ref{fig:probe-accuracy}. This performance gap between local (100\%) and global (60.6\%) probes indicates that while the symmetry information is preserved in every orbit, its orientation is not globally aligned. This suggests a `twisted' geometry where the basis of the regular representation varies depending on the input $x$.

These results confirm that the attenuated signal is not just numerical noise, but a distinguishing feature that persists despite the training objective. It underscores our core insight: the network does not structurally transition to the trivial representation; rather, it maintains the full algebraic regular structure, instantiating the target geometry solely through geometric contraction.

\section{Exploratory experiments}\label{Appendix:exploratory-exps}
\textcolor{black}{This section is organized as follows}: 
\begin{itemize}
    \item \textcolor{black}{Section \ref{appendix:extract-orbits} describes how we extract the embedded orbits and check their linear independence to get the number of linearly independent orbits.}
    \item \textcolor{black}{Section \ref{appendix:further-details-exploration} contains further details for the exploratory experiments, including hyperparameters and regularization terms for the algebra loss.}
    \item \textcolor{black}{Section \ref{appendix:piecewise-analytic-exp} repeats the TMNIST and MNIST experiments from Section \ref{sec:empirical-exploration} but for non-analytic encoders.}
    \item \textcolor{black}{Section \ref{appendix:exploratory-different-depths} repeats the TMNIST experiment from Section \ref{sec:empirical-exploration} by varying the depth of the layer considered as $Z$.}
    \item \textcolor{black}{Section \ref{appendix:tmnist-initialisations} repeats the TMNIST experiment from Section \ref{sec:empirical-exploration} by changing the initialization scheme for the learnable group action $\widehat\rho_Z$.}
\end{itemize}

\subsection{\textcolor{black}{Extracting embedded orbits and checking their linear independence}}\label{appendix:extract-orbits}
\textcolor{black}{We describe how we extract the embedded orbits and how we compute their linear independence. Let $E:\mathcal X\to Z$ denote the encoder, and $G=\{g_1,\dots,g_n\}$ the finite group considered. First, we compute the embedded orbit as $E(\mathcal O_x)=\{\widehat\rho_Z(g_i)(E(x))\}_{i\in\{1,\dots,n\}}\subseteq Z$ with $|E(\mathcal O_x)|=|G|$. Then, given embedded orbits $E(\mathcal O_{x_1}),\dots,E(\mathcal O_{x_m})$, we collect all vectors in their union in a matrix $K \in \mathbb R^{m|G|\times d}$. These orbits are linearly independent if the matrix $K$ is full rank, which is computed by checking that all its singular values are non-zero.}

\textcolor{black}{Each number in the `Orbits' columns in the Tables from Sections~\ref{sec:empirical-exploration} and Appendix \ref{appendix:piecewise-analytic-exp}, \ref{appendix:exploratory-different-depths} and \ref{appendix:tmnist-initialisations} is the maximum number of linearly independent orbits found by randomly sampling combinations of training samples $x\in \mathcal X$. For each run, we sample 500 different combinations.}

\subsection{Further details for the exploratory experiments}\label{appendix:further-details-exploration}
Here we give details of the exploratory experiments we describe in Section~\ref{sec:optimal-rep}. These use the TMNIST, MNIST and CIFAR10 datasets to determine the optimal representation on the latent space. Sections \ref{sec:tmnist-appendix}, \ref{sec:mnist-appendix} and \ref{sec:cifar-appendix} provide details of the architectures and regularisation terms used for each of these experiments. 
In all runs, we use the Adam optimiser~\cite{kingma-adam} with default parameters $(\beta_1,\beta_2)=(0.9, 0.999)$, and report additional hyperparameters in Table \ref{tab:hyperparams-optimal-rep}. These were chosen through a manual tuning process.

\subsubsection{TMNIST autoencoder, $G=C_2$}\label{sec:tmnist-appendix}
\begin{figure*}[t]
\def\mysize{2.5cm}
\[
\begin{array}{cccc}
(i)\,
\begin{aligned}
\includegraphics[width=\mysize]{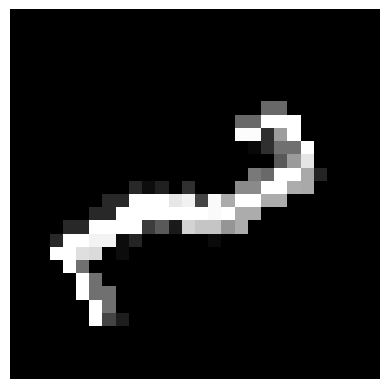}
\end{aligned}
&
(ii)\,
\begin{aligned}
\includegraphics[width=\mysize]{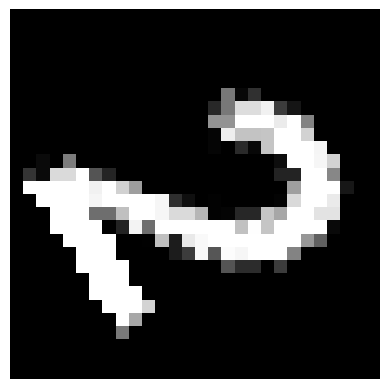}
\end{aligned}
&
(iii)\,
\begin{aligned}
\includegraphics[width=\mysize]{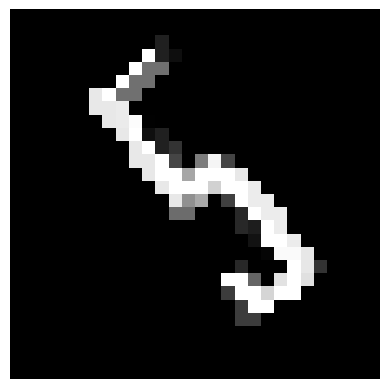}
\end{aligned}
&
(iv)\,
\begin{aligned}
\includegraphics[width=\mysize]{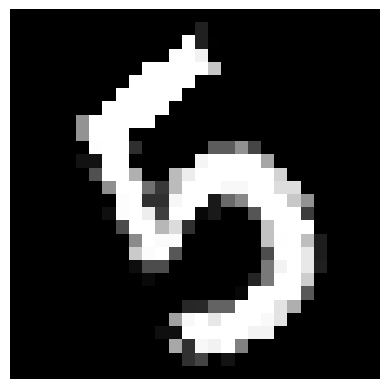}
\end{aligned}
\\
\end{array}
\]

\caption{Examples of our augmented training dataset for the TMNIST experiment, from the  chosen fonts `Bahianita-Regular' $(i)$, $(iii)$ and `IBMPlexSans-MediumItalic' $(ii)$, $(iv)$. \label{fig:tmnist-examples}}
\end{figure*}

This experiment uses the TMNIST dataset \cite{magre2022typographymnisttmnistmniststyleimage} of digits rendered in a variety of typefaces. We select a data subset corresponding to just two typefaces `IBMPlexSans-MediumItalic' and `Bahianita-Regular', and augment with 180° rotations. We give some examples of our augmented dataset in Figure~\ref{fig:tmnist-examples}. The group we use here  is $C_2=\{1,a \,|\, a^2 =1\}$ and, for a data point $x$, we define the group action $\rho_{\mathcal X}(a)(x)$ to be the data point with the font swapped, but the rotation and scaling unchanged. In particular, with reference to images Figure~\ref{fig:tmnist-examples}(i)--(iv), we have $\rho_{\mathcal X}(a)(i)=(ii)$, $\rho_{\mathcal X}(a)(ii)=(i)$, $\rho_{\mathcal X}(a)(iii)=(iv)$ and $\rho_{\mathcal X}(a)(iv) = (iii)$.
For this experiment we set $L_{\mathrm{task}}=\MSE$, and we use a simple CNN autoencoder with hyperparameters given in Table~\ref{tab:hyperparams-optimal-rep}. The architectural details can be found on the provided repository.
\begin{table}[h]

    \caption{Hyperparameters for exploratory experiments.}
\small
\setlength{\tabcolsep}{1mm}

  \centering
  \begin{tabular}{lcccccc}
    \toprule
    Experiment & Latent dim.\ & $\lambda_{a}$ & $\lambda_{t}$ & $\lambda_{e}$ & LR & Batch Size \\
    \midrule
    TMNIST $C_2$     & \textcolor{black}{8}                  & 1.0           & 0.5         & 1         & 0.003                  & 64                  \\
    MNIST $D_3$     & 18                  & 0.5           & 0.495         & 0.005         & 0.003                  & 64                  \\
    CIFAR10 $C_4$     & 16                  & 1.0           & 25         & 0.25         & 0.003                  & 64                  \\
    \bottomrule
  \end{tabular}

  \label{tab:hyperparams-optimal-rep}
\end{table}

We use the following regularization term:
\begin{equation}
    \text{REG}_{C_2, d} = \MSE( \widehat \rho_{Z}(a), \widehat \rho_{Z}(a)^{-1} )
\end{equation}
Here $\widehat \rho_{Z}(a)^{-1}$ is computed with 
$\widehat \rho_{Z}(a)^{-1} = \mathrm{\texttt{torch.linalg.solve}}(\widehat \rho_{Z}(a), \mathrm{I}_d)$ for efficiency and numerical stability. We found empirically that this regularization helps to stabilize the training of $\widehat \rho _{Z}(a)$, allowing us to achieve lower values for the algebra loss.

\subsubsection{MNIST autoencoder, $G=D_3$}\label{sec:mnist-appendix}
This experiment uses the MNIST dataset \cite{deng2012mnist} of handwritten digits. The group considered is $D_3=\{e,r,r^2,r^3,s,rs\,|\,r^3=e, s^2=e, rsrs=e\}$, and on the input space we define the group action such that  $\rho_{\mathcal X}(r)(x)$ is the counterclockwise rotation of $x$  by \textcolor{black}{120} degrees, and $\rho_{\mathcal X}(s)(x)$ is the image generated by flipping $x$ about the vertical axis. For this experiment, we set $L_{\mathrm{task}}=\MSE$, and use a simple MLP autoencoder with hyperparameters given in Table \ref{tab:hyperparams-optimal-rep}. The architectural details can be found on the provided repository.

We use the following regularization term:
\begin{equation}
    \text{REG}_{D_3, d}=-0.995\;\MSE(\widehat{\rho}_{Z}(r)\widehat{\rho}_{Z}(s)\widehat{\rho}_{Z}(r) \widehat{\rho}_{Z}(s), \mathrm{I}_d)
\end{equation}
We determined empirically that this regularization dampens the interaction between the matrices  $\widehat{\rho}_{Z}(r)$ and $\widehat{\rho}_{Z}(s)$ in a way that improves training. Low final values of the algebra loss reported in Table \ref{tab:tmnist} give evidence that we still obtain a high-quality representation despite this damping.

\subsubsection{CIFAR10 classifier, $G=C_4$}\label{sec:cifar-appendix}
This experiment uses the CIFAR10 dataset \cite{Krizhevsky2009LearningML} of 32x32 images organised in 10 classes: airplane, automobile, bird, cat, deer, dog, frog, horse, ship, truck. The group considered is the cyclic group of size four
$C_4$ of addition on the set $\{0, 1, 2, 3\}$ modulo 4. The element 1 is a generator for this group, and for an input vector $x$, we define the group action such that $\rho_{\mathcal X}(1)(x)$ is the rotation of $x$ by 90 degrees counterclockwise. For this experiment  we set $L_{\mathrm{task}}=\mathrm{CrossEntropy}$, and use a simple CNN classifier with hyperparameters given in Table \ref{tab:hyperparams-optimal-rep}. The architectural details can be found on the provided repository.

The regularization term used is the following:
\begin{equation}
    \text{REG}_{C_4, d} = \MSE( \widehat \rho_{Z}(1)^3, \widehat \rho_{Z}(1)^{-1} )
\end{equation}
Here, $\widehat \rho_{Z}(1)^{-1}$ is computed with 
$\widehat \rho_{Z}(1)^{-1} = \mathrm{\texttt{torch.linalg.solve}}(\widehat \rho_{Z}(1), \mathrm{I}_d)$ for efficiency and numerical stability. We determined empirically that this regularization helps to stabilize the training of $\widehat \rho_{Z}(1)$ and the behavior of its inverse.

\subsection{\textcolor{black}{Exploratory experiments for non-analytic encoders}}\label{appendix:piecewise-analytic-exp}
\textcolor{black}{In this section, we repeat the same experiments for TMNIST $C_2$ and MNIST $D_3$ from Section \ref{sec:empirical-exploration} but for non-analytic encoders. In particular, we use the same architecture but we replace the $\tanh$ activation function with $\text{ReLU}$. Table \ref{tab:piecewise-tmnist} shows similar results as the fully analytic encoders (Table \ref{tab:tmnist}), suggesting empirically that the optimization process avoids any potentially degenerate regions.}

\begin{table}[h]
\def\m{\text{-}}
\small
\setlength{\tabcolsep}{1mm}
\caption{Piece-wise analytic encoder experiments. Left, TMNIST autoencoder task, learned representations of $C_2$  on  latent space. Right, MNIST autoencoder task, learned representations of $D_3$ on latent space.\label{tab:piecewise-tmnist}}
    \centering
    \hspace{-1cm}
    \begin{tabular}{cccccc}
    \toprule
    &\multicolumn{2}{c}{Irrep. counts}
    \\
    \cmidrule{2-3}
        Run & \hspace{5pt}$-1$ & $+1$ & Alg.\ loss & Eq. loss & Orbs.
        \\ \midrule
        1 & \hspace{5pt}4 & 4 & $5.7 \newtimes 10^{\m 5}$ & $4.6 \newtimes 10^{\m 3}$ & 4 \\
        2 & \hspace{5pt}3 & 5 & $6.7 \newtimes 10^{\m 9}$ & $6.6 \newtimes 10^{\m 6}$ & 3\\
        3 & \hspace{5pt}4 & 4 & $2.7 \newtimes 10^{\m 8}$ & $2.5 \newtimes 10^{\m 5}$ & 4\\
        4 & \hspace{5pt}4 & 4 & $2.3 \newtimes 10^{\m 9}$ & $4.2 \newtimes 10^{\m 6}$ & 4\\
        5 & \hspace{5pt}3 & 5 & $6.0 \newtimes 10^{\m 9}$ & $1.9 \newtimes 10^{\m 5}$ & 3
        \\ \bottomrule
    \end{tabular}
    \hspace{5pt}
    \begin{tabular}{ccccccc}
        
    \toprule
    &\multicolumn{3}{c}{Irrep. counts}
    \\
    \cmidrule{2-4}
        Run & Triv & Sgn & Std & Alg.\ loss & Eq.\ loss & Orbs.
        \\ \midrule
        1 & 3.01 & 3.01 & 5.99 & $1.2\newtimes 10^{\m 3}$ & $1.3\newtimes 10^{\m 2}$ & 3 \\
        2 & 2.98 & 2.98 & 6.01 & $6.1\newtimes 10^{\m 4}$ & $2.3\newtimes 10^{\m 2}$ & 3 \\
        3 & 3.32 & 3.36 & 5.66 & $3.1\newtimes 10^{\m 2}$ & $1.4\newtimes 10^{\m 2}$ & 3 \\
        4 & 3.03 & 3.31 & 5.69 & $1.4\newtimes 10^{\m 2}$ & $1.2\newtimes 10^{\m 2}$ & 3 \\
        5 & 2.98 & 2.98 & 6.02 & $8.5\newtimes 10^{\m 4}$ & $1.3\newtimes 10^{\m 2}$ & 3 \\ \bottomrule
    \end{tabular}
    \hspace{-1cm}
\end{table}

\subsection{\textcolor{black}{Exploratory experiments at different layer depths}}\label{appendix:exploratory-different-depths}
\textcolor{black}{In this section, we repeat the TMNIST experiment from Section \ref{sec:empirical-exploration} for different layer depths. In the original experiment, we choose to study equivariance with respect to the layer $Z$ chosen as the central hidden layer (the output layer of the encoder). Table~\ref{tab:tmnist-different-depths} shows the results of choosing $Z$ as the first or last hidden layer. 
The results are similar to those in Section \ref{sec:empirical-exploration}: each linearly independent embedded orbit corresponds to a copy of the regular representation, and the network tends to learn a multiple of it.}
\begin{table}[h]
\def\m{\text{-}}
\small
\setlength{\tabcolsep}{1mm}
\caption{\textcolor{black}{TMNIST experiment with $Z$ at different depths. Left: $Z$ is taken as the first hidden layer; Right: $Z$ is taken as the final hidden layer.}\label{tab:tmnist-different-depths}}
    \centering
    \hspace{-1cm}
    \begin{tabular}{cccccc}
    \toprule
    &\multicolumn{2}{c}{Irrep. counts}
    \\
    \cmidrule{2-3}
        Run & \hspace{5pt}$-1$ & $+1$ & Alg.\ loss & Eq. loss & Orbs.
        \\ \midrule
        1 & \hspace{5pt}3 & 5 & $4.9 \newtimes 10^{\m 10}$ & $1.1 \newtimes 10^{\m 4}$ & 3 \\
        2 & \hspace{5pt}3 & 5 & $4.2 \newtimes 10^{\m 9\pz}$ & $1.6 \newtimes 10^{\m 4}$ & 3\\
        3 & \hspace{5pt}4 & 4 & $1.0 \newtimes 10^{\m 10}$ & $6.2 \newtimes 10^{\m 5}$ & 4\\
        4 & \hspace{5pt}4 & 4 & $2.3 \newtimes 10^{\m 6\pz}$ & $3.0 \newtimes 10^{\m 5}$ & 4\\
        5 & \hspace{5pt}3 & 5 & $9.0 \newtimes 10^{\m 10}$ & $1.2 \newtimes 10^{\m 4}$ & 3
        \\ \bottomrule
    \end{tabular}
    \hspace{5pt}
    \begin{tabular}{cccccc}
    \toprule
    &\multicolumn{2}{c}{Irrep. counts}
    \\
    \cmidrule{2-3}
        Run & \hspace{5pt}$-1$ & $+1$ & Alg.\ loss & Eq. loss & Orbs.
        \\ \midrule
        1 & \hspace{5pt}3 & 5 & $6.8 \newtimes 10^{\m 9\pz}$ & $4.5 \newtimes 10^{\m 4}$ & 3 \\
        2 & \hspace{5pt}4 & 4 & $9.3 \newtimes 10^{\m 10}$ & $3.7 \newtimes 10^{\m 4}$ & 4\\
        3 & \hspace{5pt}3 & 5 & $1.8 \newtimes 10^{\m 8\pz}$ & $4.5 \newtimes 10^{\m 4}$ & 3\\
        4 & \hspace{5pt}4 & 4 & $6.9 \newtimes 10^{\m 10}$ & $4.6 \newtimes 10^{\m 4}$ & 4\\
        5 & \hspace{5pt}4 & 4 & $2.3 \newtimes 10^{\m 8\pz}$ & $4.4 \newtimes 10^{\m 4}$ & 4
        \\ \bottomrule
    \end{tabular}
    \hspace{-1cm}
\end{table}

\subsection{\textcolor{black}{Exploratory experiment with different initialization}}\label{appendix:tmnist-initialisations}
\textcolor{black}{In this section, we repeat the TMNIST experiment from Section \ref{sec:empirical-exploration} but with a different initialization scheme. While Table \ref{tab:tmnist} shows results for $\widehat\rho_Z$ initialized according to a normal distribution $\mathcal N(\bm 0, \bm I_d)$, Table \ref{tab:tmnist-identitiy-init} shows results for the same experiment with $\widehat\rho_Z$ initialized close to the identity as $\bm I_d+\mathcal N(\bm 0, \bm I_d)$.}

\textcolor{black}{The results confirm Theorem \ref{thm:reg-rep}, as each linearly independent embedded orbit contributes one copy of the regular representation. However, the network typically does not learn a representation that consists entirely of a multiple of the regular representation. We observe that the trivial representation, corresponding to the eigenvalue $+1$ of $\widehat\rho_Z$ is over-represented. We hypothesize that the strong priming given by the initialization prevents a full exploration of the parameter space.
To establish the practical advantage of the regular representation, we provide ablations with the trivial representation in controlled settings (Sections \ref{sec:invariant_classification} and \ref{sec:medmnist}).}

\begin{table}[h]
\def\m{\text{-}}
\small
\setlength{\tabcolsep}{1mm}
\caption{\textcolor{black}{TMNIST experiment with $\widehat\rho_Z$ initialized close to the identity. }\label{tab:tmnist-identitiy-init}}
    \centering
    \hspace{-1cm}
    \begin{tabular}{cccccc}
    \toprule
    &\multicolumn{2}{c}{Irrep. counts}
    \\
    \cmidrule{2-3}
        Run & \hspace{5pt}$-1$ & $+1$ & Alg.\ loss & Eq. loss & Orbs.
        \\ \midrule
        1 & \hspace{5pt}2 & 6 & $3.1 \newtimes 10^{\m 5}$ & $2.9 \newtimes 10^{\m 4}$ & 2 \\
        2 & \hspace{5pt}2 & 6 & $9.5 \newtimes 10^{\m 4}$ & $0.3 \newtimes 10^{\m 4}$ & 2\\
        3 & \hspace{5pt}2& 6 & $9.5 \newtimes 10^{\m 4}$ & $1.6 \newtimes 10^{\m 4}$ & 2\\
        4 & \hspace{5pt}2 & 6 & $4.6 \newtimes 10^{\m 5}$ & $1.8 \newtimes 10^{\m 4}$ & 2\\
        5 & \hspace{5pt}2 & 6 & $2.4 \newtimes 10^{\m 5}$ & $9.1 \newtimes 10^{\m 5}$ & 2
        \\ \bottomrule
    \end{tabular}
\end{table}

\section{Main experiments}\label{appendix:main-exp}

Here we give details of the main experiments described in Section~\ref{sec:experiments}, which test our model of Section~\ref{sec:method} on tasks using the DDMNIST, MedMNIST, SMOKE and SHREC`11 datasets. 
\textcolor{black}{Section \ref{appendix:cohen-stat} discusses Cohen's $d$-statistic, which we use to assess the effect size of our intervention.} Sections \ref{sec:ddmnist-appendix}, \ref{sec:medmnist-appendix}, \ref{sec:smoke-appendix} and \ref{Appendix:shrec} provide details of the datasets, architectures and hyperparameters that we use, \textcolor{black}{together with an effect size analysis}. In all runs we use the Adam optimizer~\cite{kingma-adam} with default parameters $(\beta_1,\beta_2)=(0.9, 0.999)$, with weight decay set to 0 for DDMNIST and MedMNIST, and set to $4 \times 10^{-4}$ for SMOKE.

\color{black}
\subsection{Cohen's $d$-statistic}\label{appendix:cohen-stat}

Cohen's $d$-statistic is a widely-adopted metric \citep{miranda2025pretrainingtrulybettermetalearning, apathetic, robust-and-trustowrthy, karandikar2021soft,hermann2024analysis} to assess effect size, i.e.~the meaningfulness of the difference between distributions.
In particular, Cohen's $d$ quantifies the difference between two distributions in standard deviation units. Commonly used thresholds in machine learning are the following \citep{hermann2024analysis}:
\begin{itemize}
    \item $|d| < 0.5$, small effect
    \item $0.5 \leq |d| <0.8$, medium effect
    \item $0.8\leq |d| <1.2$, large effect
    \item $1.2\leq |d|$, very large effect
\end{itemize}
Suppose we are given $n_1$ and $n_2$ observations of two distributions, with means $\overline{x}_1$ and $\overline{x}_2$, and standard deviations $s_1$ and $s_2$ respectively. Cohen's $d$ is then defined as follows:
\begin{gather}
    d = \frac{\overline{x_1} -\overline x_2}{s},\qquad s=\sqrt\frac{(n_1-1)s_1^2+(n_2-1)s_2^2}{n_1+n_2-2}
\end{gather}
To assess the effect size of our model, we choose $\overline{x}_1$ to be the mean result of our model on a particular task, and $\overline{x}_2$ to be the mean result of a benchmark model. When reported in the tables below, we choose the sign of the effect value so that a positive value indicates our model performed better.

\color{black}

\subsection{DDMNIST experiments}
\label{sec:ddmnist-appendix}

\begin{figure}[h]
\def\mysize{2.5cm}
\[
\begin{array}{cc}
\includegraphics[width=\mysize]{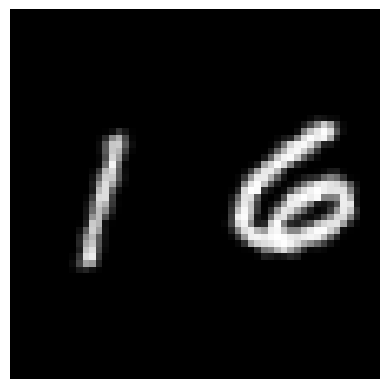}
&
\includegraphics[width=\mysize]{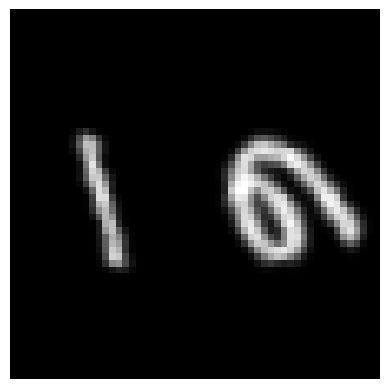}
\\
\text{Before\ augmentation}
&\text{After\ augmentation}
\end{array}
\]

\caption{Examples of training data for the DDMNIST experiment with $G=D_4$. The left figure shows concatenated MNIST digits, and the right figure shows the result after a random augmentation. In this instance, the left digit is augmented with a reflection about the vertical axis, and the right digit is augmented with a clockwise 90-degree rotation.\label{fig:ddmnist-examples}}
\end{figure}

\paragraph{Data preparation.}
We follow closely the setup of  the originators Veefkind and Cesa~\cite{Veefkind2024}. 
To generate this dataset, pairs of MNIST 28x28 images are chosen uniformly at random, and independently augmented according
to the corresponding group action for $G\in\{C_4, C_2, D_4\}$ as per Table \ref{tab:ddmnist-groups}. We give an example in Figure~\ref{fig:ddmnist-examples}. To ensure comparability of our results with the original paper, for $G \in \{C_4, D_4\}$ we follow their method of introducing interpolation artefacts by rotating each digit image by a random angle $\theta\in[0,2\pi)$, and then rotating it back by $-\theta$; for $G = C_2$ these interpolation artefacts are not added, in line with the original paper. Finally, the two images are concatenated horizontally, and padded so that the final image is $56\times56$. In this way, we obtain a dataset of 10,000 images with labels in the set $\{(0,0), (0,1), \ldots,(9,9)\}$.
\begin{table}[h]
  \caption{Symmetry groups and their actions on DDMNIST.\label{tab:ddmnist-groups}}
  \small
  \setlength{\tabcolsep}{3pt}
  \centering
  \begin{tabular}{ c  l  p{3.5cm}  c }
    \toprule
    Group & Type    & Generators                             & Size \\
    \midrule
    $C_4$ & Cyclic  & 90° rotation                            & 4 \\
    $C_2$ & Dihedral& Horizontal reflection                   & 2 \\
    $D_4$ & Dihedral& \parbox[t]{3.5cm}{Horizontal reflection\\and 90° rotation} & 8 \\
    \bottomrule
  \end{tabular}

\end{table}

\paragraph{Architecture.}
We use the same CNN architecture as in Veefkind and Cesa~\cite{Veefkind2024}, except that the final convolutional layer has an increased number of filters, from 48 to 66. We make this change  so that we can fit a copy of the regular representation of $D_4\times D_4$. 
To ensure a fair comparison, the results reported in Table \ref{tab:ddmnist}, including those for SCNN, RPP, etc, are those obtained with the increased number of filters, which we found marginally improved performance. Furthermore, we use a different learning rate for the  CNN model, as we found that this increased performance and ensured a more meaningful baseline comparison. The CNN architectural details can be found on the provided repository.

\paragraph{Hyperparameters.}
We report the hyperparameters used for the CNN and our model for the DDMNIST experiments in Table \ref{tab:ddmnist-hparams}. These hyperparameters were chosen after a grid search with the following values: learning rate~$\in \{0.001, 0.005, 0.0001, 0.0005, 0.00001, 0.00005\}$, 
and equivariance coupling strength $\lambda \in \{0.5, 1, 1.5, 2\}$. All other hyperparameters match those used by Veefkind and Cesa.
\begin{table}[h]
\caption{Hyperparameters for DDMNIST experiments.\label{tab:ddmnist-hparams}}
\setlength{\tabcolsep}{1mm}
\small
\centering
\begin{tabular}{l@{}c@{}cc@{}c@{}cc@{}c@{}cc}
\toprule
\multirow{2}{*}{}
&\,\,\,\,\,& \multicolumn{2}{c}{$C_4$}
&\,\,\,\,\,& \multicolumn{2}{c}{$C_2$}
&\,\,\,\,\,& \multicolumn{2}{c}{$D_4$} \\
    \cmidrule{3-4}
    \cmidrule{6-7}
    \cmidrule{9-10}
&& LR & $\lambda$ && LR & $\lambda$ && LR & $\lambda$ \\
\midrule
CNN && 0.0005  & - && 0.001 & - && 0.0005 & - \\
Standard rep && - & - && - & - && 0.0005 & 1 \\
Ours (regular) && 0.001 & 2 && 0.001 & 1 && 0.0005 & 1 \\
\bottomrule
\end{tabular}

\end{table}

\color{black}
\paragraph{Effect size analysis.} We report the effect size of our intervention in Table \ref{tab:ddmnist-effect-size}. For each model, the `Effect' column reports the Cohen $d$-value, comparing that model against `Ours' with the regular representation. We observe that, for each model considered, there is at least one task where the difference with our model is very large according to Cohen's $d$ statistic (Appendix \ref{appendix:cohen-stat}).

\begin{table*}[h]
\small
\setlength{\tabcolsep}{1mm}
\caption{\textcolor{black}{DDMNIST test accuracies and effect sizes. Mean over 3 runs; standard deviation in brackets. Best result in each column is bold, second-best is underlined. For $C_2, C_4$ the defining representation is equivalent to the regular representation and so is omitted. Effect values compare to `Ours (regular)', and a positive value means ours performed better. The annotations *, **, *** indicate medium, large and very large effect sizes respectively.}}\label{tab:ddmnist-effect-size}
\centering
    \begin{tabular}{l@{\hspace{10mm}}cc@{\hspace{10mm}}cc@{\hspace{10mm}}cc}
    \toprule
    {Model}
     &
     {$C_4 \uparrow$}
     &
     {Effect}
     &
     {$C_2 \uparrow$}
     &
     {Effect}
     &
     {$D_4 \uparrow$}
     &
     Effect
    \\\midrule
        CNN    & 0.907 \sd{0.004} & \pz 2.0*** & \underline{0.938} \sd{0.006} & \pz \pz 1.8*** & 0.800 \sd{0.001} & 43.0*** \\
        SCNN   & 0.484 \sd{0.008} & 68.2*** & 0.474 \sd{0.003} & 133.8*** & 0.431 \sd{0.010} & 60.6*** \\
        Restriction & \underline{0.914} \sd{0.007} & \pz 0.2 \phantom * \phantom * & 0.890 \sd{0.007} & \pz 10.0*** & 0.837 \sd{0.020} & \pz 2.2*** \\
        RPP    & 0.908 \sd{0.022} & \pz 0.4 \phantom * \phantom * & 0.903 \sd{0.009} & \pz \pz 6.3*** & 0.827 \sd{0.020} & \pz 2.9*** \\
        PSCNN  & 0.909 \sd{0.007} & \pz 1.1**\phantom * & 0.871 \sd{0.016} & \pz \pz 6.5*** & \underline{0.842} \sd{0.011} & \pz 3.3*** \\ \midrule
  Trivial rep        & 0.874 \sd{0.004} & 10.0*** & 0.938 \sd{0.007} & \pz \pz 1.6*** & 0.819 \sd{0.004} & 15.5*** \\
  Defining rep       & \qquad -- &  & \qquad -- &  & 0.838 \sd{0.010} & \pz 4.2*** \\
Ours (regular)      & \textbf{0.915} \sd{0.004} &  & \textbf{0.947} \sd{0.004} &  & \textbf{0.868} \sd{0.002} &  \\
\bottomrule
    \end{tabular}

\end{table*}

\color{black}



\subsection{MedMNIST experiments}
\label{sec:medmnist-appendix}

\paragraph{Data preparation.}
For this experiment, we use three subsets of the MedMNIST dataset~\cite{Yang_2023}, in line with Veefkind and Cesa~\cite{Veefkind2024}: Nodule3D, Synapse3D and Organ3D, each containing 3D images of size 28x28x28.
Nodule3D is a public lung nodule dataset, containing 3D images from thoracic CT scans; for this dataset, the task is to classify each nodule as benign or malignant. Synapse3D contains 3D images obtained from an adult rat with a multi-beam scanning electron microscope; the task is to classify whether a synapse is excitatory or inhibitory. Organ3D is a classification task for a 3D images of human body organs, with the following labels: liver, right kidney, left kidney, right femur, left femur, bladder, heart, right lung, left lung, spleen and pancreas.

For augmentations, we choose the octahedral group of orientation-preserving rotational symmetries of the cube, which is isomorphic to the permutation group $S_4$. We define its action $\rho_{\mathcal X}(g)$ on a 3D image $x$ by applying the corresponding rotational symmetry of the cube. Specifically, we parameterise $g$ as a tuple $(l, \theta)$ where $l=(x,y,z)$ specifies a rotation axis and $\theta$ specifies the rotation angle about the axis $l$ to obtain 24 rotation matrices each with size $3\times3$, one for each of the 24 elements of $S_4$. In summary, we have  rotation matrices corresponding to the following tuples:

\begin{description}[
    font=\normalfont,          
    labelwidth=3.5em,          
    labelsep=0.5em,            
    leftmargin=! ,             
    itemsep=2pt,               
    parsep=0pt                 
]
  \item[Identity (1)]  
    $(l,0)$ for any $l$.
  \item[Coord-axis (9)]  
    $(l,\theta)$ for 
    $l\in\{(1,0,0),(0,1,0),(0,0,1)\}$ and 
    $\theta\in\{\tfrac\pi2,\pi,\tfrac{3\pi}2\}$.
  \item[Edge-mid (6)]  
    $(l,\theta)$ for 
    $l\in\{(1,1,0),(1,-1,0),(1,0,1),$ 
    $(1,0,-1),(0,1,1),(0,1,-1)\}$ \\and 
    $\theta=\pi$.
  \item[Body-diag (8)]  
    $(l,\theta)$ for 
    $l\in\{(1,1,1),(1,1,-1),$\\$(1,-1,1),(-1,1,1)\}$ and 
    $\theta\in\{\tfrac{2\pi}3,\tfrac{4\pi}3\}$.
\end{description}


\paragraph{Architecture.}
For these experiments we use the same CNN-based ResNet architecture as Veefkind and Cesa~\cite{Veefkind2024}. This is formed from seven 3D convolutional layers, formed into 3 blocks with residual connections, along with batch normalisation and pooling. The architectural details can be found on the provided repository.

\paragraph{Hyperparameters.}We report the hyperparameters used for the baseline with $S_4$ augmentations, and for our model in the MedMNIST experiments in Table \ref{tab:medmnist-hparams}. These hyperparameters were chosen after a grid search with the following values: learning rate~$\in \{0.001, 0.005, 0.0001, 0.0005, 0.00001, 0.00005\}$, and equivariance coupling strength $\lambda \in \{0.5, 1, 1.5, 2\}$. All other hyperparameters are the same as those used by Veefkind and Cesa.%
\begin{table}[h]

\caption{Hyperparameters for MedMNIST experiments.\label{tab:medmnist-hparams}}
\small
\setlength{\tabcolsep}{1mm}
\centering
\begin{tabular}{l@{}c@{}cc@{}c@{}cc@{}c@{}cc}
\toprule
\multirow{2}{*}{}
&\,\,\,\,\,& \multicolumn{2}{c}{Nodule3D}
&\,\,\,\,\,& \multicolumn{2}{c}{Synapse3D}
&\,\,\,\,\,& \multicolumn{2}{c}{Organ3D} \\
    \cmidrule{3-4}
    \cmidrule{6-7}
    \cmidrule{9-10}
&& LR & $\lambda$ && LR & $\lambda$ && LR & $\lambda$ \\
\midrule
CNN (Augmented) && 0.00005 & - && 0.0001 & - && 0.0001 & - \\
Ours && 0.00005 & 1 && 0.0001 & 1 && 0.0001 & 2 \\
\bottomrule
\end{tabular}

\end{table}

\color{black}
\paragraph{Effect size analysis.}
We report the effect size of our intervention in Table \ref{tab:medmnist-effect}. For each model, the `Effect' column reports the Cohen $d$-value comparing that model against `Ours' with the regular representation. We observe that, for each model considered, there is at least one task where the difference with our model is very large according to Cohen's $d$ statistic (Appendix \ref{appendix:cohen-stat}).
\color{black}

\begin{table}[h]
\small
\setlength{\tabcolsep}{1mm}
\caption{\textcolor{black}{MedMNIST3D test accuracies and effect sizes. Mean over 3 runs; standard deviation in brackets. Parameter counts shown. Best result in each column is bold, second-best is underlined. Effect values compare to `Ours (regular)', and a positive value means ours performed better. The annotations *, **, *** indicate medium, large and very large effect sizes respectively.\label{tab:medmnist-effect}}}
\centering
    \begin{tabular}{llcc@{\hspace{6mm}}cc@{\hspace{6mm}}cc}
\toprule
Group & Model  & Nodule$\,\uparrow$ & Effect & Synapse$\,\uparrow$ & Effect & Organ$\,\uparrow$ & Effect
\\ \midrule
        N/A & CNN  & 0.873 \sd{0.005} & 2.80*** & 0.716 \sd{0.008} & 9.26*** & 0.920 \sd{0.003} & -7.01*** \\
        Aug & CNN   & \underline{0.879} \sd{0.007} & 1.32*** & 0.761 \sd{0.008} & 1.54*** & 0.632 \sd{0.005} & 0.25 {\phantom *} {} \\
        SO(3) & SCNN  & 0.873 \sd{0.002} & 3.68*** & 0.738 \sd{0.009} & 4.91*** & 0.607 \sd{0.006} & 0.88**  \\
        SO(3) & RPP  & 0.801 \sd{0.003} & 20.86*** & 0.695 \sd{0.037} & 2.86*** & \underline{0.936} \sd{0.002} & -7.42*** \\
        SO(3) & PSCNN  & 0.871 \sd{0.001} & 4.44*** & \textbf{0.770} \sd{0.030} & 0.00 \phantom * \phantom * & 0.902 \sd{0.006} & -6.53***\\
        O(3) & SCNN  & 0.868 \sd{0.009} & 2.61*** & 0.743 \sd{0.004} & 8.54*** & 0.902 \sd{0.006} & -6.53*** \\
        O(3) & RPP  & 0.810 \sd{0.013} & 7.82*** & 0.722 \sd{0.023} & 2.94*** & \textbf{0.940} \sd{0.006} & -7.48***\\
        O(3) &PSCNN  & 0.873 \sd{0.008} & 2.10*** & \underline{0.769} \sd{0.005} & 0.26 \phantom * \phantom * & 0.905 \sd{0.004} & -6.62*** \\  \midrule
        $\mathrm{Sym}_{\mathrm{cube}}$ & Trivial rep  & 0.867 \sd{0.001} & 5.55*** & 0.743 \sd{0.002} & 13.50*** & 0.571 \sd{0.002} & 1.79*** \\
        
        $\mathrm{Sym}_{\mathrm{cube}}$ & Defining rep  & 0.837 \sd{0.013} & 5.08*** & 0.756 \sd{0.019} & 1.04** & 0.560 \sd{0.025} & 1.89*** \\
        $\mathrm{Sym}_{\mathrm{cube}}$ & Ours (regular)  & \textbf{0.887} \sd{0.005} &  & \textbf{0.770} \sd{0.002} &  & 0.642 \sd{0.056} &  \\
        \bottomrule
    \end{tabular}

\end{table}

\color{black}

\subsection{SMOKE experiment}
\label{sec:smoke-appendix}

\begin{figure}[h]
  \centering
  \setlength{\abovecaptionskip}{4pt}
  \setlength{\belowcaptionskip}{4pt}
  \includegraphics[width=0.25\columnwidth]{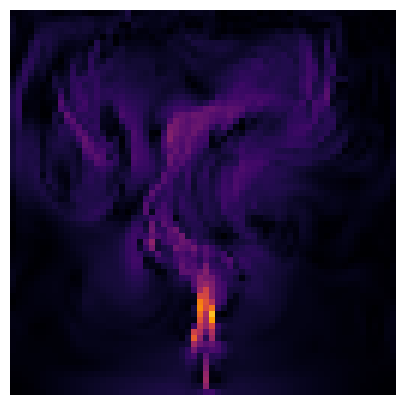}
  \includegraphics[width=0.25\columnwidth]{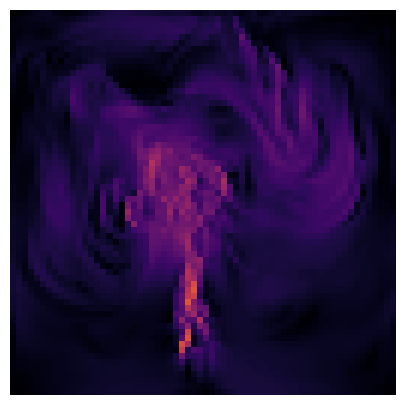}
  \caption{Approximately equivariant dynamics of smoke plumes~\cite{holl2020phiflow}.}
  \label{fig:smokevis}
\end{figure}

\begin{figure}[h]
  \centering
  \setlength{\tabcolsep}{2pt}
  \def\imgw{0.2\columnwidth}
  \begin{tabular}{@{}ccc@{}}
    \includegraphics[width=\imgw]{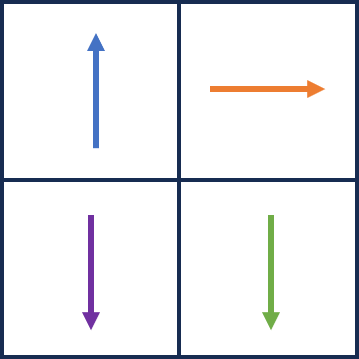} &
    \includegraphics[width=\imgw]{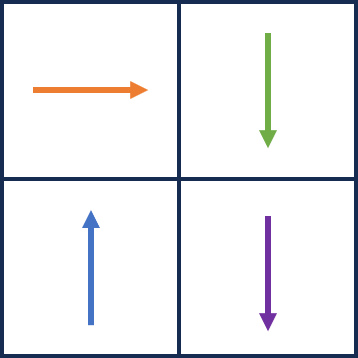} &
    \includegraphics[width=\imgw]{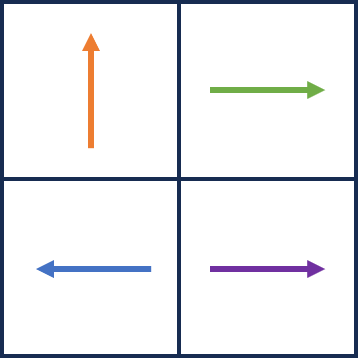} \\[2pt]
    \shortstack[c]{Original\\velocity field} 
    & \shortstack[c]{Rotation without\\reorientation} 
    & \shortstack[c]{Rotation with\\reorientation} 
  \end{tabular}
  \setlength{\abovecaptionskip}{4pt}
  \setlength{\belowcaptionskip}{4pt}
  \caption{Examples of a velocity field and its augmentations with and without reorientation. Rotating by 90° counterclockwise without reorienting simply moves the spatial grid, but breaks the physical meaning of the underlying system.}
  \label{fig:reorientation-example}
\end{figure}

\paragraph{Data preparation.}
Here we use the SMOKE dataset of Wang et al.~\cite{wang2022approximately}, which consists of smoke simulations with varying intial conditions and external forces presented as grids of $(x,y)$ components of a velocity field (see Figure~\ref{fig:smokevis} for a visualisation). The task is to  predict the next 6 frames of the simulation given the first 10 frames only. We evaluate each model on two metrics: Future, where the test set contains future extensions of instances in the training set; and Domain, where the test and training sets are from different instances. \textcolor{black}{In line with \cite{wang2022approximately}, we consider the group $C_4$ acting on the data by 90° rotations and reorientation of the velocity field, as illustrated in Figure \ref{fig:reorientation-example}.}

\paragraph{Architecture.}
We use the same CNN architecture, train and evaluation setups as in Veefkind and Cesa~\cite{Veefkind2024}, which they reproduced from Wang et al.~\cite{wang2022approximately}. The architectural details can be found on the provided repository. Because the latent space has the same geometric structure as the input data, i.e. ${ Z}=\mathbb{R}^c\times\mathbb{R}^h\times\mathbb{R}^w$ (channels$\times$height$\times$width), we choose a representation of $C_4$ given by the regular representation in each channel separately.

\paragraph{Hyperparameters.}
For both CNN models, with $C_4$ augmentations and without, and for our model, we use a learning rate of 0.001. Additionally, for our model, we set $\lambda=0.005$.  These hyperparameters were chosen after a grid search with the following values: learning rate~$\in \{0.001, 0.005, 0.0001, 0.0005\}$, and equivariance coupling strength $\lambda \in \{0.005, 0.05, 0.5, 1\}$. For all other hyperparameters, we copy the values used by Veefkind and Cesa.

\color{black}
\paragraph{Effect size analysis.}
We report the effect size of our intervention in Table \ref{tab:smoke-effect}. For each model, the `Effect' column reports the Cohen $d$-value comparing that model against `Ours' with the regular representation. We observe that, for each model considered, there is at least one metric where the difference with our model is very large according to Cohen's $d$ statistic (Appendix \ref{appendix:cohen-stat}).

\begin{table}[h]
\small
\setlength{\tabcolsep}{1mm}
\centering
\caption{\textcolor{black}{Test RMSE and effect for the \smoke\ dataset. Effect values compare to `Ours', and a positive value means ours performed better. The annotations *, **, *** indicate medium, large and very large effect sizes respectively.}}\label{tab:smoke-effect}
\begin{tabular}{l l @{\hspace{6mm}} c c @{\hspace{6mm}} c c}
\toprule
Group & Model & Future$\,\downarrow$ & Effect & Domain$\,\downarrow$ & Effect\\
\midrule
N/A & CNN      & 0.81 \sd{0.01} & 3.0*** & 0.63 \sd{0.00} & 2.8*** \\
Aug & CNN      & 0.83 \sd{0.03} & 2.2*** & 0.67 \sd{0.06} & 1.4*** \\
N/A & MLP      & 1.38 \sd{0.06} & 14.0*** & 1.34 \sd{0.03} & 32.6*** \\
C4  & E2CNN    & 1.05 \sd{0.06} & 6.3*** & 0.76 \sd{0.02} & 9.5*** \\
C4  & RPP      & 0.96 \sd{0.10} & 2.5*** & 0.82 \sd{0.01} & 21.0*** \\
C4  & Lift     & 0.82 \sd{0.01} & 4.0*** & 0.73 \sd{0.02} & 7.6*** \\
C4  & RGroup   & 0.82 \sd{0.01} & 4.0*** & 0.73 \sd{0.02} & 7.6*** \\
C4  & RSteer   & 0.80 \sd{0.00} & 2.8*** & 0.67 \sd{0.01} & 6.0*** \\
C4  & PSCNN    & \textbf{0.77} \sd{0.01} & -1.0** & \textbf{0.57} \sd{0.00} & -5.7*** \\
\midrule
C4  & Ours     & \underline{0.78} \sd{0.01} &  & \underline{0.61} \sd{0.01} &  \\
\bottomrule
\end{tabular}

\end{table}

\color{black}

\subsection{SHREC `11 experiment}\label{Appendix:shrec}
\paragraph{Data preparation.} We use the SHREC `11 dataset~\cite{lianSHREC11TrackShape2011, mitchelMobiusConvolutionsSpherical2022} where each 3D shape is also transformed with conformal transformations. We perform augmentation according to the group $O_h$ of octahedral symmetries.

\paragraph{Architecture.} We use the same architecture as the original authors~\cite{mitchelNeuralIsometriesTaming2024}, which is a ResNet-based autoencoder. Similarly to the smoke experiment, the latent space retains a geometric structure. Therefore, we choose a representation of $O_h$ given by the regular representation in each channel separately. 

\paragraph{Hyperparameters.} Due to computational constraints, we do not perform hyperparameter tuning, and we keep the same hyperparameters as the original authors~\cite{mitchelNeuralIsometriesTaming2024}, except that we set the batch size to 4. We set $\lambda=0.5$. Additionally, we symmetrize the equivariance loss to the decoder too, i.e., with $\lambda'=0.8$,
\begin{equation*}
    \lambda'\;|| \rho_{\mathcal X}(g)(x) - D(\rho_{Z}(g)(E(x)))||
\end{equation*}

\color{black}
\paragraph{Effect size analysis.}
We report the effect size of our intervention in Table \ref{tab:shrec-effect}. For each model, the `Effect' column reports the Cohen $d$-value comparing that model against `Ours' with the regular representation. We observe that, for the augmented baseline and NFT, the difference with our model is very large according to Cohen's $d$ statistic (Appendix \ref{appendix:cohen-stat}). The same analysis reveals that NIso and our model are essentially equivalent on this task.
\color{black}

\begin{table}[h]
\small
\setlength{\tabcolsep}{1mm}
\centering
\caption{\textcolor{black}{Test accuracies and effect for the SHREC '11 dataset. Effect values compare to `Ours', and a positive value means ours performed better. The annotations *, **, *** indicate medium, large and very large effect sizes respectively.}}\label{tab:shrec-effect}

\begin{tabular}{l@{\hspace{2mm}} c c} \\
\toprule
Model & Acc.$\,\uparrow$ & Effect\\
\midrule
NIso \cite{mitchelNeuralIsometriesTaming2024} & \underline{90.26} \sd{1.27} & 0.1 \\
NFT \cite{koyamaNeuralFourierTransform2024a} & 83.24 \sd{2.03} & 3.5*** \\
AE with aug & 69.36 \sd{2.81} & 8.5*** \\
MC \cite{mitchelMobiusConvolutionsSpherical2022} & 86.5 & -- \\
\midrule
Ours & \textbf{90.45} \sd{2.1} &  \\
\bottomrule
\end{tabular}

\end{table}

\newpage
\section{Sensitivity Analysis}\label{Appendix:sensitivity}
\begin{figure}[h]
    \centering
    \caption{Mean accuracy and standard deviation (over 5 runs) for different values of $\lambda$ on the DDMNIST $D_4$ task. $\lambda=0$ is the baseline.}
    \includegraphics[width=0.5\linewidth]{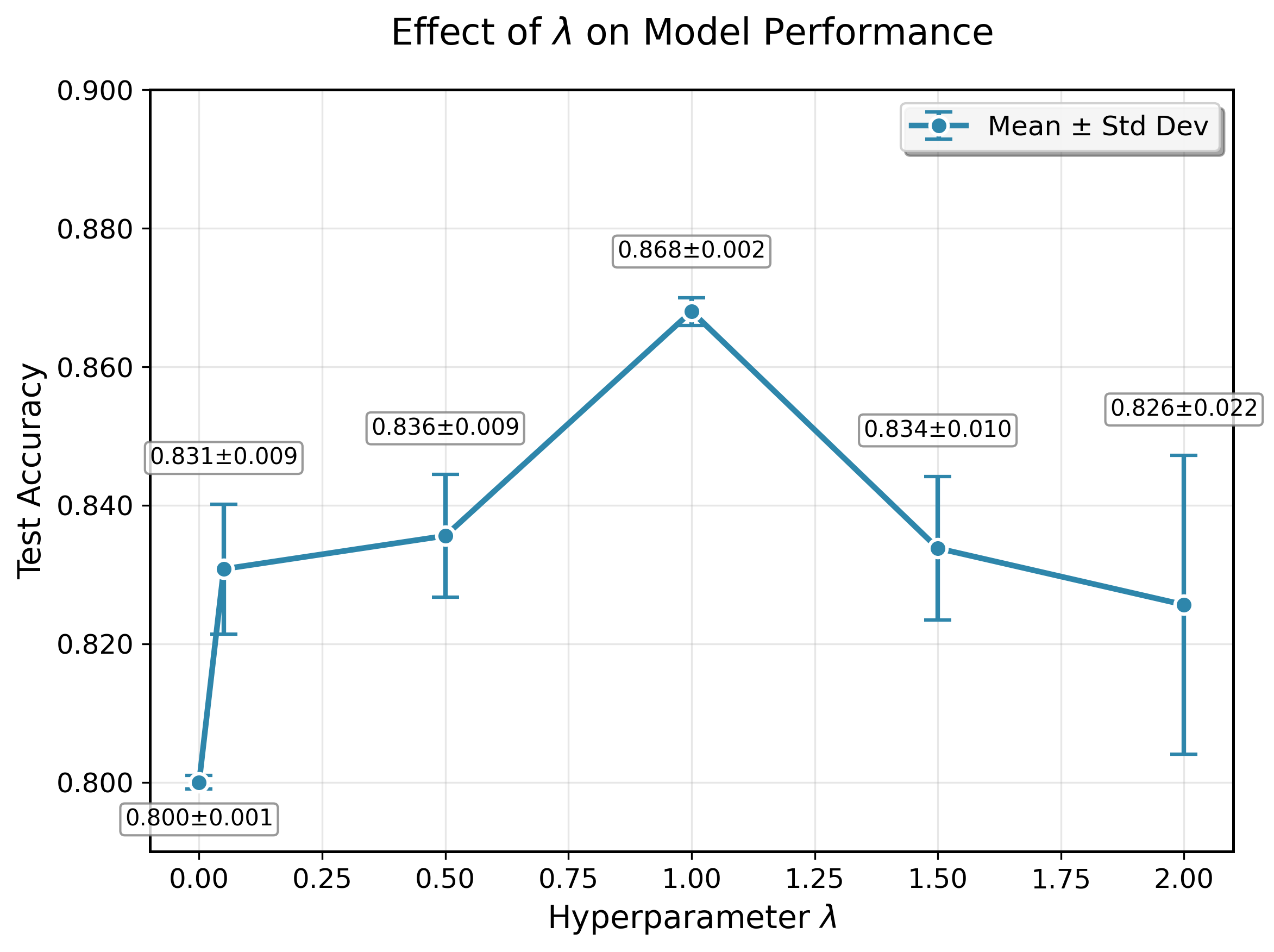}
    \label{fig:sensitivity}
\end{figure}

To assess the practical usability of our method, we performed a sensitivity analysis on the hyperparameter $\lambda$, which controls the strength of the equivariance loss. We evaluated our model on the DDMNIST $D_4$ task across six different values for $\lambda$: $\{0,0.05, 0.5, 1, 1.5, 2\}$, with $\lambda=0$ being the baseline. The results, reported in Figure~\ref{fig:sensitivity}, show that while peak performance is achieved at $\lambda=1$, the model maintains high accuracy and low variance across a wide range of values (0.5 to 2.0). This analysis demonstrates that our method is robust to the specific choice of $\lambda$.

\end{document}